\documentclass[10pt,twocolumn,letterpaper]{article}

\usepackage[pagenumbers]{cvpr} 

\usepackage[table,xcdraw]{xcolor}
\usepackage{booktabs}
\usepackage{xspace}
\usepackage{array}    
\usepackage{amsmath}  

\renewcommand{\arraystretch}{1.05}

\newcommand{\method}{\textsc{LAM3C}\xspace}
\newcommand{\data}{\texttt{RoomTours}\xspace}

\newlength\savewidth
\usepackage{multirow}

\newcommand\tablestyle[2]{%
  \setlength{\tabcolsep}{#1}%
  \renewcommand{\arraystretch}{#2}%
}

\newcolumntype{x}[1]{>{\centering\arraybackslash}p{#1pt}}
\newcolumntype{y}[1]{>{\raggedleft\arraybackslash}p{#1pt}}
\newcolumntype{z}[1]{>{\raggedright\arraybackslash}p{#1pt}}

\newcommand\shline{%
  \noalign{\vskip 3pt}%
  \specialrule{.1em}{.05em}{.05em}%
  \noalign{\vskip 3pt}%
}

\usepackage{pifont}
\newcommand{\cmark}{\ding{51}}

\definecolor{baselinecolor}{gray}{.9}
\newcommand\baseline[1]{\cellcolor{baselinecolor}{#1}}

\definecolor{cvprblue}{rgb}{0.21,0.49,0.74}
\usepackage[pagebackref,breaklinks,colorlinks,allcolors=cvprblue]{hyperref}

\definecolor{cvprblue}{rgb}{0.21,0.49,0.74}
\usepackage[pagebackref,breaklinks,colorlinks,allcolors=cvprblue]{hyperref}

\def\paperID{*****} 
\def\confName{CVPR}
\def\confYear{2026}

\title{
3D \textit{sans} 3D Scans: Scalable Pre-training from Video-Generated Point Clouds
}

\author{Ryousuke Yamada\textsuperscript{1,2} \qquad
Kohsuke Ide \textsuperscript{1} \qquad
Yoshihiro Fukuhara \textsuperscript{1} \qquad
Hirokatsu Kataoka\textsuperscript{1,3} \\[3mm]
Gilles Puy \textsuperscript{4,\thanks{indicates Valeo.ai.}} \qquad
Andrei Bursuc \textsuperscript{4,\footnotemark[1]} \qquad
Yuki M. Asano \textsuperscript{2} \\[3mm]
\textsuperscript{1} AIST \qquad \textsuperscript{2} University of Technology Nuremberg \qquad \textsuperscript{3} University of Oxford \qquad \textsuperscript{4} INRIA \vspace{-2mm}
}

\begin{document}
\maketitle
\begin{abstract}
Despite recent progress in 3D self-supervised learning, collecting large-scale 3D scene scans remains expensive and labor-intensive. In this work, we investigate whether 3D representations can be learned from unlabeled videos recorded without any real 3D sensors. We present Laplacian-Aware Multi-level 3D Clustering with Sinkhorn-Knopp (\method{}), a self-supervised framework that learns from video-generated point clouds reconstructed from unlabeled videos. We first introduce \data, a video-generated point cloud dataset constructed by collecting room-walkthrough videos from the web (e.g., real-estate tours) and generating 49,219 scenes using an off-the-shelf feed-forward reconstruction model.  We also propose a noise-regularized loss that stabilizes representation learning by enforcing local geometric smoothness and ensuring feature stability under noisy point clouds. Remarkably, without using any real 3D scans, \method{} achieves better performance than previous self-supervised methods on indoor semantic and instance segmentation. These results suggest that unlabeled videos represent an abundant source of data for 3D self-supervised learning. Our source code is available at \url{https://github.com/ryosuke-yamada/lam3c}.
\end{abstract}    
\section{Introduction}
\label{sec:intro}
Understanding real-world 3D scenes is essential for AI systems that require visual spatial intelligence~\cite{yang2025thinking}. As of 2025, vision foundation models (VFMs) such as DINOv2/v3~\cite{oquab2023dinov2, simeoni2025dinov3} and SAM~\cite{Kirillov_2023_ICCV} have shown remarkable generalization in image recognition tasks. DINOv3 is pre-trained on about 1.7 billion unlabeled images, while SAM is trained on 11 million densely annotated images. In contrast, 3D data is highly constrained by scanning real environments and manual annotation, making large-scale collection extremely difficult. The largest widely used indoor scene dataset offers only about 5k unique scenes~\cite{baruch2021arkitscenes}. This disparity in data scalability poses significant limitations to achieving scalable 3D pre-training.

\begin{figure}[t]
  \centering
  \includegraphics[width=\linewidth]{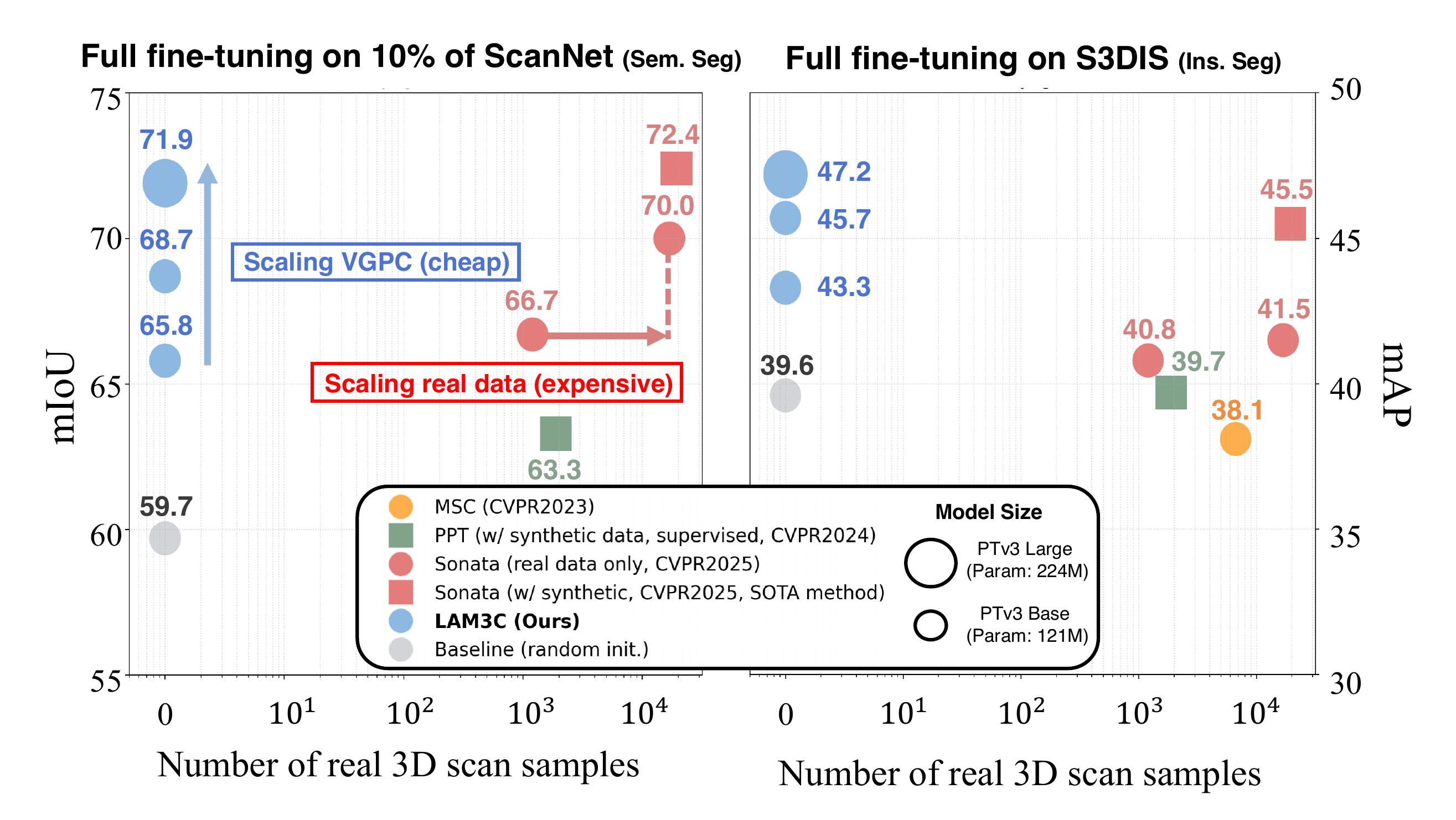}
  \vspace{-10pt}
  \caption{\textbf{Video-generated point clouds (VGPC) match or exceed real 3D scan performance without using any real 3D data.} Left: Our method (LAM3C) trained solely on VGPC achieves comparable performance to methods trained on real 3D scans when fine-tuning on 10\% of ScanNet. Right: Instance segmentation results on S3DIS show LAM3C outperforms self-supervised methods trained on real 3D scans and matches Sonata which uses both real and synthetic data.}
  \label{fig:data_efficincy}
  \vspace{-10pt}
\end{figure}

3D self-supervised learning (3D-SSL) for indoor scene understanding has made steady progress~\cite{wu2024towards,wu2023masked,xie2020pointcontrast}. For instance, Sonata~\cite{wu2025sonata} is a representative approach that extends DINOv2 to point clouds and mitigates the geometric shortcut problem, improving segmentation performance. However, most existing 3D-SSL methods remain fundamentally limited by the availability of 3D data. Even when combining real and synthetic point clouds, the training scale is around 140k samples, and merely 18k for real 3D scans alone. Such limited data scale hinders 3D-SSL methods from reaching the same level of success achieved in 2D vision. This limitation motivates us to develop a new 3D pre-training framework without relying on real 3D scans.

We hypothesize that we can extract enough geometric cues from unlabeled videos to learn 3D representations. Traditionally, computer vision has aimed to infer 3D structures from multi-view images. Classical methods such as Structure-from-Motion (SfM)~\cite{ullman1979interpretation, wu2011visualsfm, agarwal2011building, tomasi1992shape} and Multi-View Stereo (MVS)~\cite{okutomi1993multiple, furukawa2009accurate, seitz1999photorealistic} reconstruct 3D structure by identifying correspondences across multi-view images and optimizing camera parameters and point clouds. While these classical methods are accurate, they involve a complex optimization procedure. More recently, feed-forward reconstruction models such as VGGT have emerged~\cite{wang2025vggt,Wang_2024_CVPR,keetha2025mapanything}. Despite their simple design, these models can directly infer 3D structure from multi-view images and achieve reconstructions that are comparable to, or even better than, classical methods. These advances indicate that videos contain strong geometric cues that modern reconstruction models can extract. Building on this insight, we argue that this opens a path towards \textit{scan-free} indoor scene understanding.

We aim to leverage video-generated point clouds (VGPC) reconstructed from unlabeled videos as pre-training data and show that they are remarkably effective pre-training signals for 3D-SSL. We introduce Laplacian-Aware Multi-level 3D Clustering with Sinkhorn-Knopp (\method{}), a pre-training framework that learns stable representations from VGPC via  Sinkhorn-Knopp  clustering~\cite{knight2008sinkhorn,cuturi2013sinkhorn,asano2019self,caron2020unsupervised}. \method{} comprises the following key components: (i) \data. We construct a new dataset, \data, by collecting room-walkthrough videos from the web (e.g., real-estate tours) and reconstructing VGPC from various camera locations using an off-the-shelf feed-forward 3D reconstruction model. \data contains 49,219 VGPC scenes reconstructed from indoor videos.
(ii) Noise-regularized loss. VGPC contain noise and missing regions, which makes point-wise embeddings unstable during Sinkhorn-Knopp clustering. To address this, we introduce a noise-regularized loss composed of two terms: a Laplacian smoothing term, which encourages spatially adjacent points to produce similar embeddings by smoothing features along the local geometry of the VGPC; a noise consistency term, which enforces two views from the same scene to have similar global representations even in the presence of noisy regions. Together, these losses make learning robust to noisy VGPC and lead to stable representations. The main contributions of this paper are as follows.

\begin{itemize}
    \item  We propose a pipeline for generating large datasets of VGPC of indoor scenes from unlabeled videos. Thanks to this pipeline, we create \data, a dataset of 49k point clouds obtained from videos collected from the web.
    \item We propose \method{}, a self-supervised method allowing stable representation learning on imperfect point clouds such as those reconstructed from videos. We achieve this thanks to two new regularization losses used to control the effect of noise and missing regions in the point clouds.
    \item We show that VGPC are sufficient to learn 3D representations. We  outperform Sonata \cite{wu2025sonata} trained only on real 3D scans on several benchmarks (see \cref{fig:data_efficincy}). Beyond \method{}, the key ingredients are a sufficiently large dataset of VGPC, like \data, a large-capacity 3D backbone, and sufficiently long pre-training schedule. 
\end{itemize}

\section{Related work}
\label{sec:formatting}

\noindent{\textbf{SSL on point clouds.}}
Manual annotation of real 3D scans is costly, making 3D-SSL, which generates pseudo-supervision from unlabeled data, an essential technique for representation learning. Early 3D-SSL methods concentrated on building representations for dense scans of single objects~\cite{chen2021shape,poursaeed2020self,sauder2019self}. In indoor scene understanding, PointContrast~\cite{xie2020pointcontrast} pioneered 3D-SSL by performing point-level contrastive learning, establishing the first successful pre-training on large-scale point clouds. Subsequent works improved performance thanks to a better choice of contrasting pairs~\cite{depthcontrast,strl,liu2023fac,wu2023masked,groupcontrast,Hou_2021_CVPR}, and also extended the application to sparse point clouds of large outdoor scenes~\cite{proposalcontrast,segcontrast,tarl,bevcontrast}. Beyond contrastive methods, other strategies used a reconstruction pretext task~\cite{voxelmae1,voxelmae2,pointmae,pointm2ae,pointbert,also}. Recently, inspired by the success of DINO~\cite{caron2021emerging,oquab2023dinov2}, Sonata  adapted this technique to point clouds with recipes to mitigate low-level geometric shortcuts during pre-training. It achieved notable gains over the state-of-the-art, especially in linear probing.  

Despite these advances, progress in 3D-SSL remains limited by the quantity of data available. Even with mixed real and synthetic data, Sonata’s largest setup involves only about 140k scenes, constrained by the high cost and complexity of acquiring real 3D scans. Without addressing this bottleneck, the progress of 3D-SSL may remain fundamentally limited by data scale. Our work departs from this setting by leveraging unlabeled videos instead of real 3D scans, providing a scalable alternative for representation learning.

\noindent{\textbf{Feed-forward reconstruction models.}}
Feed-forward reconstruction models have emerged, marking a significant shift in the paradigm of 3D reconstruction~\cite{Wang_2024_CVPR, Cabon_2025_CVPR,wang2025vggt, keetha2025mapanything}. VGGT~\cite{wang2025vggt} jointly predicts camera poses and point maps from multi-view images using a single transformer model, eliminating the need for sequential SfM optimization~\cite{okutomi1993multiple, furukawa2009accurate, seitz1999photorealistic}. In addition, $\pi^3$~\cite{wang2025pi} introduces a fully permutation-equivariant architecture, enabling affine-invariant camera poses and scale-invariant local point maps without a reference view that is invariant to the ordering of input views. This design yields a more robust reconstruction that is well suited to in-the-wild domains. While these models achieve high-quality reconstruction without explicit geometric optimization, they only focus on reconstruction itself rather than representation learning for 3D scene understanding. The potential of the intermediate features or reconstructed point clouds for pre-training has not yet been fully realized.

This paper explores the potential of VGPC, produced by a feed-forward reconstruction model, as pre-training data for understanding indoor scenes. Our proposed \method{} achieves representation learning without real 3D scans.

\section{\method{}}

\begin{figure}[t]
  \centering
  \includegraphics[width=\linewidth]{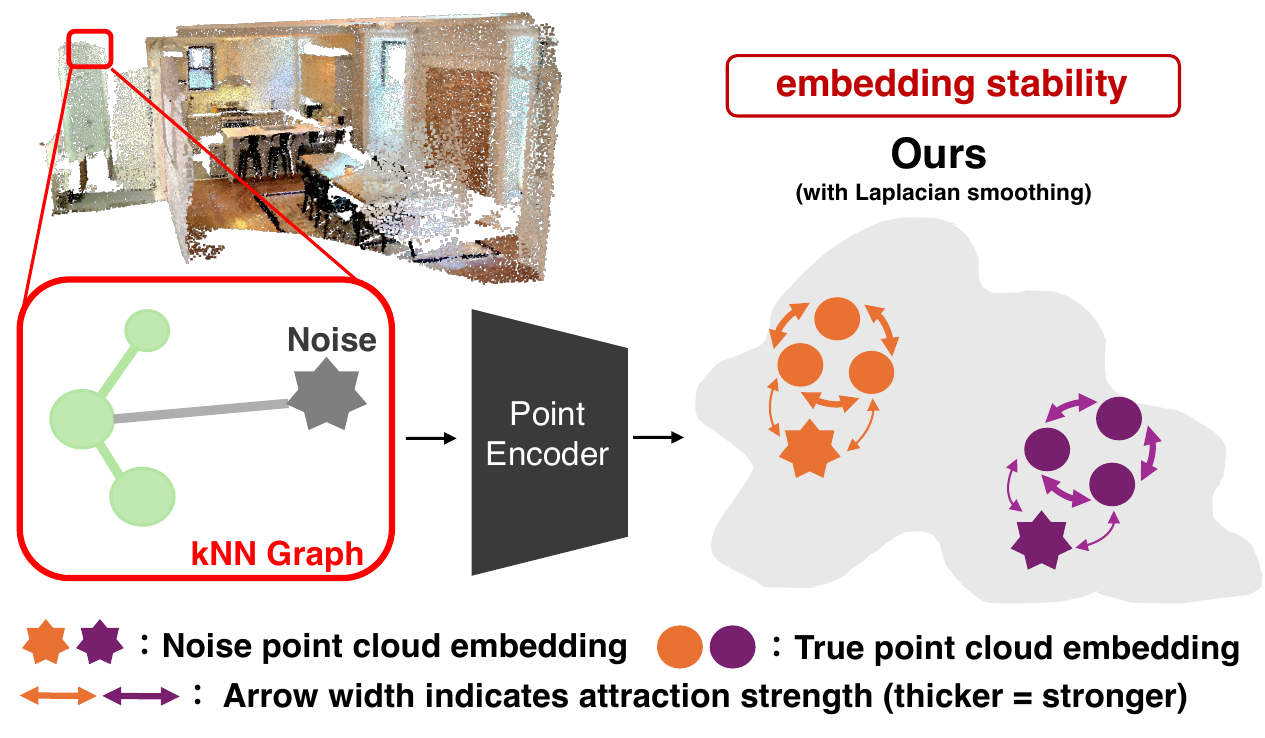}
  \caption{\textbf{Laplacian smoothing loss.} \method{} resolves learning instability by using kNN graphs for VGPC and assigning distance-based weights. This reduces embedding divergence caused by noisy point clouds and promotes more stable representations from neighboring point clouds.
  }
  \label{fig:laplacian}
  \vspace{-10pt}
\end{figure}

This section introduces our proposed \method{} as a pre-training framework. 
\method{} learns 3D representations through Sinkhorn-Knopp clustering from VGPC generated by a feed-forward reconstruction model.
The architecture follows a teacher–student design, where the student model is trained to mimic the feature representations produced by the teacher. The teacher model is updated using an exponential moving average (EMA)~\cite{polyak1992acceleration} of the student parameters. This design encourages more generalizable representations, whereas robustness to noisy or missing point clouds comes from our noise-regularized losses. 
In addition to the standard clustering loss~\cite{wu2025sonata}, \method{} introduces two new losses that mitigate the noisy nature of VGPC to enhance local smoothness and feature consistency. We recall the Sonata loss  and describe our new noise-regularized losses below.

\noindent{\textbf{Base clustering loss.}}
We build upon the clustering loss, which has shown strong performance for scene understanding. The teacher and student models are fed with different augmentations of the same scene, and the teacher’s parameters are updated as the EMA of the student’s parameters to stabilize training. The overall clustering loss is given by:
\begin{equation}
\mathcal{L}_{\text{clustering}} =
w_u \mathcal{L}_{\text{unmask}} +
w_m \mathcal{L}_{\text{mask}} +
w_r \mathcal{L}_{\text{roll}}.
\end{equation}

The unmask loss $\mathcal{L}_{\text{unmask}}$ aligns student local-view features to teacher global-view features via kNN matching. The mask loss $\mathcal{L}_{\text{mask}}$ distills the teacher’s prototype assignments from a full global-view to the student on a masked global-view. The roll-mask loss $\mathcal{L}_{\text{roll}}$ swaps the two global-views when forming targets to enforce cross-view consistency. Following the Sonata configuration, we set the loss weights to $w_u,w_m,w_r=4:2:2$. Each loss term aligns the feature embeddings through a cross-entropy loss over prototype assignments:
\begin{equation}
\mathcal{L}_k = -\sum_i q_i^{(t)} \log p_i^{(s)}.
\end{equation}

Here, $p_i^{(s)} = \mathrm{softmax}(z_i^{(s)} / \tau_s)$ denotes the student’s probability distribution, and $q_i^{(t)} = \mathrm{Sinkhorn}(z_i^{(t)} / \tau_t)$ represents the teacher’s entropy-regularized soft assignment, following~\cite{asano2019self, caron2020unsupervised}. Both $z_i^{(s)}$ and $z_i^{(t)}$ are prototype logits, while $\tau_s$ and $\tau_t$ are temperature parameters that control the sharpness of the distributions.

\begin{figure}[t]
  \centering
  \includegraphics[width=\linewidth]{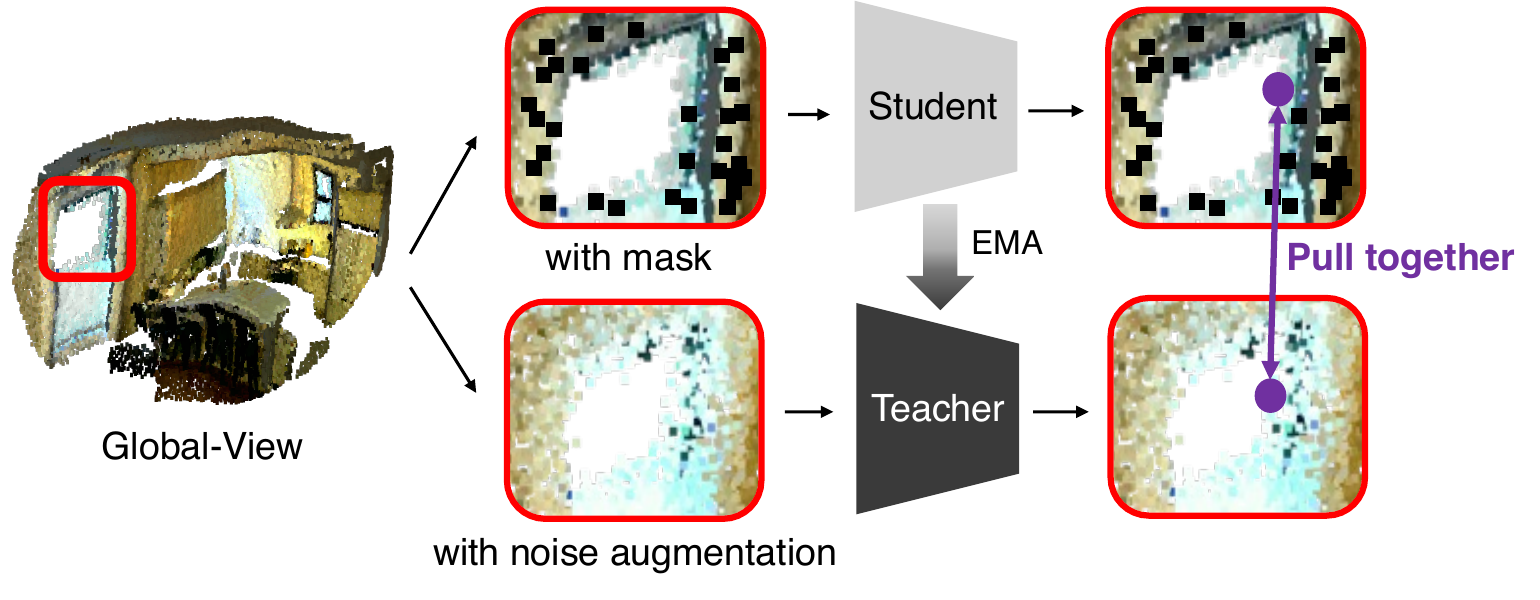}
  \caption{\textbf{Noise consistency loss.} 
  This is a constraint term stating that the same point cloud for the teacher and student models should yield the same embedding in the feature space. This enables more stable clustering even in VGPC.
  }
  \label{fig:noise_consistecy}
  \vspace{-10pt}
\end{figure}

\noindent{\textbf{Laplacian smoothing loss.}}
VGPC contain noise and missing regions due to imperfect reconstruction. To stabilize representation learning under such point clouds, we construct a k-Nearest Neighbor (kNN) graph on each point cloud and apply a Laplacian smoothness loss as shown in Fig.~\ref{fig:laplacian}. For a point \( x_i \) with embedding \( z_i = f_{\theta}(x_i) \),  the loss encourages nearby points to produce similar embeddings:
\begin{equation}
R_{\text{Lap}} = \sum_{(i,j)\in E} w_{ij} \| z_i - z_j \|^2,
\quad
\end{equation}
We assign each edge a distance-based weight,
\(
w_{ij} = \exp(-{\| p_i - p_j \|^2}/{\sigma^2}),
\)
so that closer points contribute more to the smoothness constraint. We compute the edge weight \( w_{ij} \) based on the $L2$ distance between points in the VGPC. The scale parameter \( \sigma \) is adaptively estimated per point cloud as the median of the kNN distance, making the weighting robust to variations in point density. In addition, neighbors further away than a certain threshold are removed for improved robustness.
In practice, we compute the regularization on a kNN graph with distance-based edge weights and degree normalization. To improve robustness to noisy reconstructions and outliers, we replace the squared $L2$ edge penalty with a Huber penalty.

This loss smooths features along the local geometry by encouraging nearby points to share similar embeddings, while noisy point clouds receive small weights. Because the loss depends only on the relational structure between points rather than absolute coordinates, it prevents the collapse of learning and promotes stable representations.

\noindent{\textbf{Noise consistency loss.}}
While Laplacian smoothing enforces geometric smoothness across neighboring points, our noise consistency loss ensures feature stability for different augmented views of the same point cloud. As shown in Fig.~\ref{fig:noise_consistecy}, given two views of the same VGPC $x^{(a)}$ and $x^{(b)}$, we feed them into the EMA teacher $g_{\text{EMA}}$ and the student $f_\theta$, respectively, and minimize the discrepancy between their embeddings:

\begin{equation}
R_{\text{cons}}
= \frac{1}{|\mathcal{P}|}
  \sum_{(i,j)\in\mathcal{P}}
  \left\| g_{\text{EMA}} (x^{(a)})_{j}
        - f_{\theta} (x^{(b)})_{i}
  \right\|^{2},
\end{equation}
where $\mathcal{P}$ denotes the set of kNN correspondences between the noise-augmented global-view $x^{(a)}$ and the masked global-view $x^{(b)}$. Each pair $(i,j)\in\mathcal{P}$ represents corresponding points from the student and teacher inputs, respectively. $g_{\text{EMA}}(\cdot)$ and $f_\theta(\cdot)$ denote the backbone encoders that output point-wise embeddings. Minimizing $R_{\text{cons}}$ enforces that the same point maintains a consistent embedding even in the presence of noise.

\noindent{\textbf{Total objective.}}
 \method{}  extends the clustering objective by adding Laplacian smoothness and noise consistency. The final loss is defined as:
\begin{equation}
\mathcal{L}_{\text{total}}
= \mathcal{L}_{\text{clustering}}
+ \lambda R_{\text{Lap}}
+ \mu R_{\text{cons}}.
\label{eq:loss}
\end{equation}

During pre-training, we schedule the strength of regularization. The Laplacian coefficient $\lambda$ starts at $2\text{e}{-4}$ and linearly increases to $3\text{e}{-3}$, while the noise consistency coefficient $\mu$ is fixed at $0.05$. Additional ablations on $\lambda$ and $\mu$ are provided in the supplementary material. By combining Laplacian smoothing with noise consistency, our noise-regularized clustering enables stable representation learning even from noisy point clouds.

\method{}'s loss does not rely on hand-crafted indoor priors such as object scale or topology. In $L_{\text{clustering}}$, the student predicts targets computed from global views while observing only local views or masked global views. Since the targets encode global context, the task cannot be solved using local geometry alone, even for similar local structures. This encourages the representation to capture global context without relying on indoor priors.

\section{\data{} }
\label{method:dataset}
This section describes \data and the process used to generate VGPC from unlabeled videos on the web.

\begin{figure}[t]
  \centering
  \includegraphics[width=\linewidth]{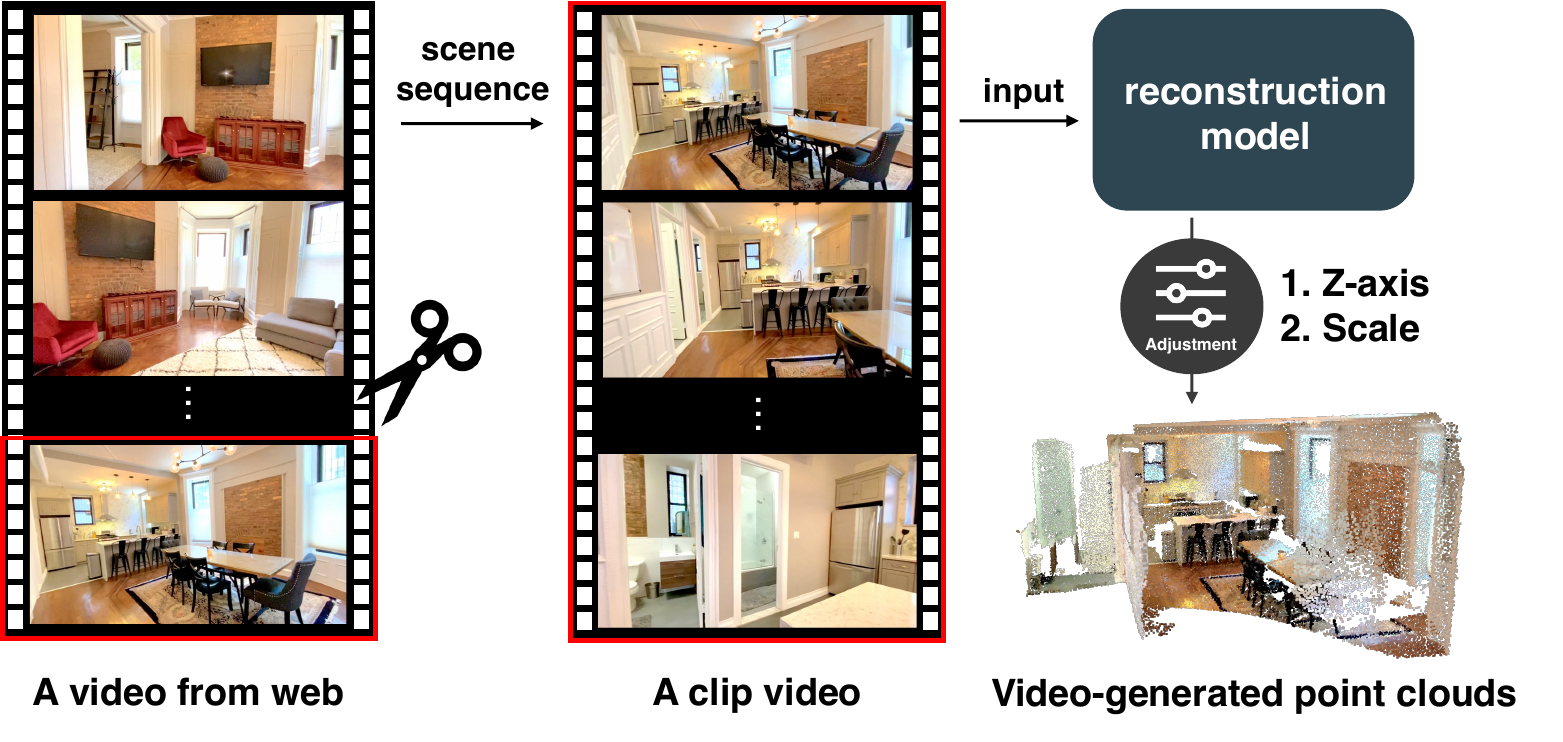}
  \caption{\textbf{Overview of \data construction.} We segment the video into scene sequences using CLIP~\cite{radford2021learning}, and generate VGPC by inputting each scene sequence into $\pi^3$. Because the scenes differ from real 3D scans in coordinate system, scale, and spacing, we apply a post-processing alignment.}
  \vspace{-10pt}
  \label{fig:roomtours_overview}
\end{figure}

\noindent{\textbf{Overview of \data{}.}}
Online video platforms such as YouTube host a vast number of indoor walkthrough videos, including real-estate tours and apartment viewings. Although these videos are unlabeled, they contain geometric cues for understanding indoor scenes, such as various scene layouts. To leverage this video resource for 3D-SSL, we construct \data, a video-generated point cloud dataset from unlabeled videos as shown in Fig.~\ref{fig:roomtours_overview}.

\noindent{\textbf{Video collection.}}
We collect indoor videos from YouTube. Since room layout and furniture vary significantly across regions in the world, we target videos from multiple cities worldwide. We search using keywords such as ``${<}\texttt{city}{>}$, real-estate, walk-through,’’ and manually select candidate channels that upload indoor tours.

To ensure accurate 3D reconstruction, we prioritize videos with fewer dynamic objects (e.g., real-estate agents walking in front of the camera) and minimal editing cuts or aggressive camera motion. Once candidate channels are selected, videos are automatically filtered based on metadata such as duration and undesired keywords (e.g., ``CG,’’ ``drone,’’ ``short’’). Only videos that satisfy these criteria are downloaded, resulting in a total of 3,462 videos. In addition, we utilize videos from RealEstate10k~\cite{zhou2018stereo}, the YouTube House Tours Dataset~\cite{chang2020semantic}, and HouseTours~\cite{Celen_2025_ICCV}.

\begin{figure*}[t]
  \centering
  \includegraphics[width=\linewidth]{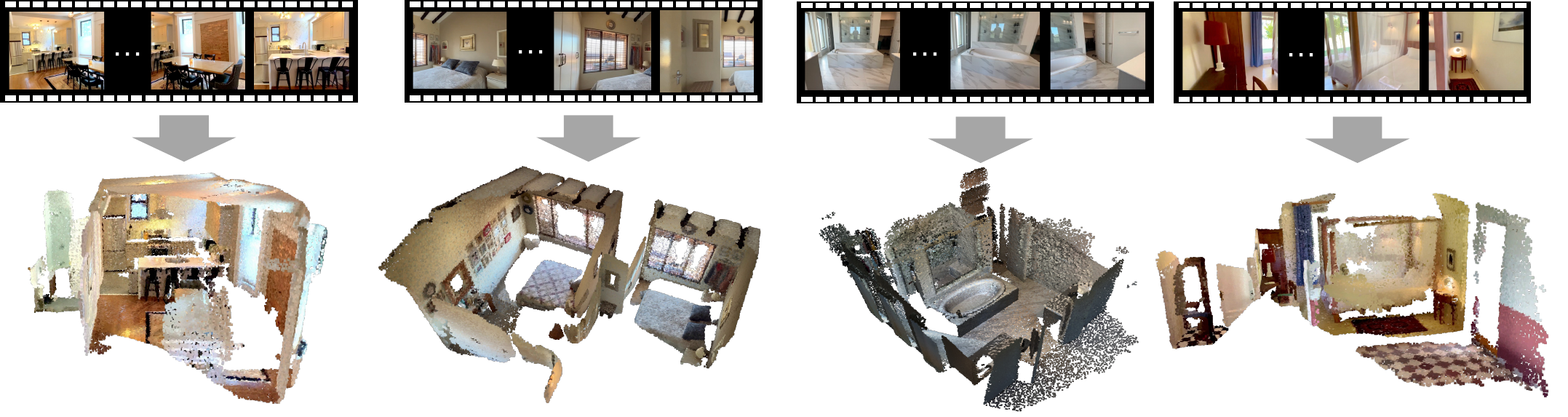}
  \caption{\textbf{Visualization of VGPC from the \data.} 
   These are pseudo-scenes generated as VGPC from unsupervised videos collected from the web. Visually, they achieve very high-quality reconstruction of real-world indoor scenes. However, for example, the leftmost scene contains a large amount of noisy point clouds in the scene due to camera shake in the input video, resulting in blurred object boundaries and cases where walls or floors appear doubled.}
  \label{fig:visualization}
  \vspace{-10pt}
\end{figure*}

\noindent{\textbf{Video segmentation for indoor scene.}}
The collected videos are not guaranteed to be indoor scenes, nor do they always correspond to a single scene.  Videos may include outdoor shots, advertisements, or multiple indoor areas from a walkthrough.  To address this issue, we convert each video into multiple indoor scene sequences.

We apply frame-level zero-shot scene classification using CLIP~\cite{radford2021learning}. In the first step, each frame is classified as indoor or outdoor, and frames predicted as outdoor are discarded. In the second step, the remaining indoor frames are grouped into three broad scene types, namely “living room”, “bedroom”, and “bathroom”.  These categories are deliberately coarse because fine-grained categories would split a video into many short sequences, making reconstruction unreliable, whereas coarser categories would mix widely different scenes into a single sequence. Scene boundaries are detected by changes in the predicted scene type, and the video is split accordingly. We stabilize the predictions by enforcing temporal consistency, ensuring the same class label for frames within 0.5 seconds.

\noindent{\textbf{VGPC generation.}}
We create VGPC from videos using $\pi^3$~\cite{wang2025pi}, which is a feed‑forward 3D reconstruction model that does not rely on SLAM~\cite{cheeseman1987stochastic,davison2007monoslam} and Bundle Adjustment~\cite{triggs1999bundle} optimization. Given a scene sequence, we first sample frames at uniform intervals and, for long sequences, automatically downsample to a target number of frames (e.g., 200-400 frames per scene). Each frame is resized to meet a per-image pixel count while preserving aspect ratio. $\pi^3$ processes the sequence in a single mixed-precision forward pass. It filters unreliable predictions using confidence masking and edge suppression,
removes remaining outliers, and then fuses the per-frame estimates into final colored point clouds. Our method outputs colored point clouds for each scene, with an average runtime of about 5 minutes. Example outputs of \data{} are shown in Fig.~\ref{fig:visualization}. 

\begin{table*}[t]
\centering
\caption{\textbf{Indoor Semantic Segmentation Results across Datasets.} All methods are evaluated by full fine-tuning (Full-FT) or linear probing (LP) on PTv3 (Base) for 100 epochs. We use mIoU as evaluation metric. VGPC indicates video-generated point clouds.}
\vspace{-10pt}
\label{tab:indoor-semseg}
\scalebox{0.97}{%
\begin{tabular}{p{28.5mm}rp{7.5mm}ccccccccccc}
\toprule
Semantic Seg. &\multicolumn{3}{c}{Pre-training Data} &\multicolumn{2}{c}{ScanNet~\cite{dai2017scannet}} &\multicolumn{2}{c}{ScanNet200~\cite{rozenberszki2022language}} &\multicolumn{2}{c}{ScanNet++ Val~\cite{yeshwanth2023scannet++}} &\multicolumn{2}{c}{S3DIS Area 5~\cite{Armeni_2016_CVPR}} \\
\cmidrule(lr){1-1} \cmidrule(lr){2-4} \cmidrule(lr){5-6} \cmidrule(lr){7-8} \cmidrule(lr){9-10} \cmidrule(lr){11-12}
Methods &\multicolumn{1}{c}{Real} &\multicolumn{1}{c}{Synth.} &\multicolumn{1}{c}{VGPC} & LP & Full-FT & LP & Full-FT & LP & Full-FT & LP & Full-FT  \\
\midrule
PTv3 & \multicolumn{1}{c}{--} & \multicolumn{1}{c}{--} & \multicolumn{1}{c}{--}
 & 16.1 & 74.7 
 & 2.2 & 32.0
 & 6.9 & 40.3
 & 29.6 & 67.8 
 \\

\midrule

 \rowcolor[gray]{0.9} \textcolor{darkgray}{MSC~\cite{wu2023masked}} & \multicolumn{1}{c}{\textcolor{darkgray}{7k}} & \multicolumn{1}{c}{\textcolor{darkgray}{--}} & \multicolumn{1}{c}{--} 
 & \textcolor{darkgray}{21.8} & \textcolor{darkgray}{78.2} 
 & \textcolor{darkgray}{3.3} & \textcolor{darkgray}{33.4} 
 & \textcolor{darkgray}{8.1} & \textcolor{darkgray}{42.4} 
 & \textcolor{darkgray}{32.1} & \textcolor{darkgray}{69.9} 
 \\

 \rowcolor[gray]{0.9} \textcolor{darkgray}{PPT~\cite{wu2024towards} (sup.)} & \multicolumn{1}{c}{\textcolor{darkgray}{1k}} & \multicolumn{1}{c}{\textcolor{darkgray}{21k}} & \multicolumn{1}{c}{--} 
 & \textcolor{darkgray}{--} & \textcolor{darkgray}{78.6} 
 & \textcolor{darkgray}{--} & \textcolor{darkgray}{36.0} 
 & \textcolor{darkgray}{--} & \textcolor{darkgray}{43.3} 
 & \textcolor{darkgray}{--} & \textcolor{darkgray}{74.3} 
 \\

 \rowcolor[gray]{0.9} \textcolor{darkgray}{Sonata~\cite{wu2025sonata}} & \multicolumn{1}{c}{\textcolor{darkgray}{18k}} & \multicolumn{1}{c}{\textcolor{darkgray}{121k}} & \multicolumn{1}{c}{--}
 & \textcolor{darkgray}{72.5} & \textcolor{darkgray}{79.4} 
 & \textcolor{darkgray}{29.3} & \textcolor{darkgray}{36.8} 
 & \textcolor{darkgray}{37.3} & \textcolor{darkgray}{43.7} 
 & \textcolor{darkgray}{72.3} & \textcolor{darkgray}{76.0} 
 \\

 \midrule

  Sonata (all real) & \multicolumn{1}{c}{15k} & \multicolumn{1}{c}{--} & \multicolumn{1}{c}{--}
 & 69.4 & 78.5 
 & 28.1 & 35.3
 & \textbf{36.3} & 42.7 
 & \textbf{69.8} & 75.2 
 \\

  Sonata (ScanNet) & \multicolumn{1}{c}{1k} & \multicolumn{1}{c}{--} & \multicolumn{1}{c}{--}
 & 67.1 & 75.4 
 & 27.2 & 32.2
 & 34.3 & 41.7 
 & 61.2 & 72.2  
 \\

  \rowcolor[HTML]{DEF3E7} \method{} (ours) & \multicolumn{1}{c}{--} &  \multicolumn{1}{c}{--} & \multicolumn{1}{c}{16k} 
 &\cellcolor[HTML]{def3e6}58.9 &\cellcolor[HTML]{def3e6}75.6
 &\cellcolor[HTML]{def3e6}23.7 &\cellcolor[HTML]{def3e6}32.8
 &\cellcolor[HTML]{def3e6}33.5 &\cellcolor[HTML]{def3e6}40.7
 &\cellcolor[HTML]{def3e6}63.8 &\cellcolor[HTML]{def3e6}71.9
 \\

  \rowcolor[HTML]{DEF3E7}  \method{} (ours) & \multicolumn{1}{c}{--} &  \multicolumn{1}{c}{--} & \multicolumn{1}{c}{49k} 
  &\cellcolor[HTML]{def3e6}66.0 &\cellcolor[HTML]{def3e6}77.7
 &\cellcolor[HTML]{def3e6}25.3 &\cellcolor[HTML]{def3e6}35.1
 &\cellcolor[HTML]{def3e6}34.2 &\cellcolor[HTML]{def3e6}43.1
 &\cellcolor[HTML]{def3e6}65.7 &\cellcolor[HTML]{def3e6}72.9
 \\
 
  \rowcolor[HTML]{DEF3E7} \method{}* (ours) & \multicolumn{1}{c}{--} & 
 \multicolumn{1}{c}{--} &  \multicolumn{1}{c}{49k} 
 & \textbf{69.5} & \textbf{79.5} 
 & \textbf{28.1} & \textbf{35.9} 
 & 35.9 & \textbf{43.1}
 & 69.5 & \textbf{75.5}
 \\

\bottomrule
\multicolumn{12}{l}{%
{* uses PTv3 (Large) and 437k pre-training steps. 
\colorbox{baselinecolor}{Gray} indicates sup.~method or access to val/test splits during pre-training.
}
}\\
\end{tabular}
}
\end{table*}

\begin{table*}[t]
\centering
\caption{\textbf{Indoor Instance Segmentation Results across Datasets.} All  methods are evaluated by full fine-tuning (Full-FT) or linear probing (LP) on PTv3 (Base) for 100 epochs. We use mAP as evaluation metric. VGPC indicates video-generated point clouds.}
\vspace{-10pt}
\label{tab:indoor-insseg}
\scalebox{0.97}{%
\begin{tabular}{p{28.5mm}rp{7.5mm}ccccccccccc}
\toprule
Instance Seg. &\multicolumn{3}{c}{Pre-training Data} &\multicolumn{2}{c}{ScanNet~\cite{dai2017scannet}} &\multicolumn{2}{c}{ScanNet200~\cite{rozenberszki2022language}} &\multicolumn{2}{c}{ScanNet++ Val~\cite{yeshwanth2023scannet++}} &\multicolumn{2}{c}{S3DIS Area 5~\cite{Armeni_2016_CVPR}} \\
\cmidrule(lr){1-1} \cmidrule(lr){2-4} \cmidrule(lr){5-6} \cmidrule(lr){7-8} \cmidrule(lr){9-10} \cmidrule(lr){11-12}
Methods &\multicolumn{1}{c}{Real} &\multicolumn{1}{c}{Synth} &\multicolumn{1}{c}{VGPC} & LP & Full-FT & LP & Full-FT & LP & Full-FT & LP & Full-FT  \\
\midrule
PTv3 & \multicolumn{1}{c}{{--}} & \multicolumn{1}{c}{{--}} & \multicolumn{1}{c}{{--}}
 & 0.2 & 26.9 
 & 0.02 & 17.2 
 & 0.03 & 18.5
 & 11.9 & 39.6 
 \\

\midrule

 \rowcolor[gray]{0.9} \textcolor{darkgray}{MSC~\cite{wu2023masked}} & \multicolumn{1}{c}{\textcolor{darkgray}{7k}} & \multicolumn{1}{c}{\textcolor{darkgray}{--}} & \multicolumn{1}{c}{{--}} 
 & \textcolor{darkgray}{--} & \textcolor{darkgray}{41.1} 
 & \textcolor{darkgray}{--} & \textcolor{darkgray}{23.4} 
 & \textcolor{darkgray}{--} & \textcolor{darkgray}{21.7} 
 & \textcolor{darkgray}{--} & \textcolor{darkgray}{38.1} 
 \\

 \rowcolor[gray]{0.9} \textcolor{darkgray}{PPT~\cite{wu2024towards}~(sup.)} & \multicolumn{1}{c}{\textcolor{darkgray}{1k}} & \multicolumn{1}{c}{\textcolor{darkgray}{21k}} & \multicolumn{1}{c}{{--}}
 & \textcolor{darkgray}{--} & \textcolor{darkgray}{42.1} 
 & \textcolor{darkgray}{--} & \textcolor{darkgray}{24.0} 
 & \textcolor{darkgray}{--} & \textcolor{darkgray}{21.9} 
 & \textcolor{darkgray}{--} & \textcolor{darkgray}{39.7} 
 \\

  \rowcolor[gray]{0.9} \textcolor{darkgray}{Sonata~\cite{wu2025sonata}} & \multicolumn{1}{c}{\textcolor{darkgray}{18k}} & \multicolumn{1}{c}{\textcolor{darkgray}{121k}} & \multicolumn{1}{c}{{--}}
 & \textcolor{darkgray}{--} & \textcolor{darkgray}{42.4} 
 & \textcolor{darkgray}{--} & \textcolor{darkgray}{25.4} 
 & \textcolor{darkgray}{--} & \textcolor{darkgray}{22.3} 
 & \textcolor{darkgray}{--} & \textcolor{darkgray}{45.5} 
 \\

 \midrule

 Sonata (all real) & \multicolumn{1}{c}{15k} & \multicolumn{1}{c}{--} & \multicolumn{1}{c}{{--}}
 & 28.0 & 40.3 
 & \textbf{9.7} & 20.8
 & 10.0 & 19.7 
 & 22.9 & 41.5  
 \\

  Sonata (ScanNet) & \multicolumn{1}{c}{1k} & \multicolumn{1}{c}{--} & \multicolumn{1}{c}{{--}}
 & 23.9 & 29.8 
 & 8.6 & 19.7
 & 9.7 & 18.7 
 & 19.5 & 40.8  
 \\

  \rowcolor[HTML]{DEF3E7} \method{} (ours) & \multicolumn{1}{c}{--} & \multicolumn{1}{c}{{--}} &  \multicolumn{1}{c}{16k}
  &\cellcolor[HTML]{def3e6}19.1 &\cellcolor[HTML]{def3e6}38.9
 &\cellcolor[HTML]{def3e6}5.8 &\cellcolor[HTML]{def3e6}18.7 
 &\cellcolor[HTML]{def3e6}9.5 &\cellcolor[HTML]{def3e6}19.7
 &\cellcolor[HTML]{def3e6}18.0 &\cellcolor[HTML]{def3e6}43.3 
 \\

   \rowcolor[HTML]{DEF3E7} \method{} (ours) & \multicolumn{1}{c}{--} & \multicolumn{1}{c}{{--}} &  \multicolumn{1}{c}{49k}
  &\cellcolor[HTML]{def3e6}25.1 &\cellcolor[HTML]{def3e6}39.7
 &\cellcolor[HTML]{def3e6}8.3 &\cellcolor[HTML]{def3e6}19.6 
 &\cellcolor[HTML]{def3e6}11.3 &\cellcolor[HTML]{def3e6}20.5
 &\cellcolor[HTML]{def3e6}21.6 &\cellcolor[HTML]{def3e6}45.7 
 \\

 \rowcolor[HTML]{DEF3E7} \method{}* (ours) & \multicolumn{1}{c}{--} & \multicolumn{1}{c}{{--}} & \multicolumn{1}{c}{49k}
 &\cellcolor[HTML]{def3e6}\textbf{28.6} &\cellcolor[HTML]{def3e6}\textbf{41.7} 
 &\cellcolor[HTML]{def3e6}9.5 &\cellcolor[HTML]{def3e6}\textbf{21.9} 
 &\cellcolor[HTML]{def3e6}\textbf{12.1} &\cellcolor[HTML]{def3e6}\textbf{21.1} 
 &\cellcolor[HTML]{def3e6}\textbf{27.8} &\cellcolor[HTML]{def3e6}\textbf{47.2}
 \\

\bottomrule
\multicolumn{12}{l}{%
* uses PTv3 (Large) and 437k pre-training steps. 
\colorbox{baselinecolor}{Gray} indicates sup.~method or access to val/test splits during pre-training.
}
\end{tabular}
}
\vspace{-10pt}
\end{table*}

\noindent{\textbf{Alignment of scenes from VGPC.}}
A scene reconstructed from $\pi^3$ often contains noise and has an inconsistent coordinate system or scale. Thus, we apply a lightweight alignment procedure that enforces geometric consistency while maintaining data diversity.
First, we randomly sample points from VGPC and remove statistical outliers using a standard SOR filter, which eliminates points with abnormally large k-NN distances. Next, we detect a dominant plane and align the scene to a Z-up coordinate system. We then perform an SVD refinement of its orientation. If plane detection fails, ceilings or walls may be mistakenly selected, causing the coordinate system to flip. We keep such scenes unchanged in our dataset because automatic rejection could remove valid samples and reduce diversity.

In addition, we define the scale of a scene as the diagonal length of its axis-aligned bounding box, which captures the overall spatial extent of the geometry. This value is denoted as \(s_{\text{current}}\).  We then align the scene with a factor $\alpha = \frac{s_{\text{target}}}{s_{\text{current}}}$, where \( s_{\text{target}} \) is drawn from a scale distribution estimated in advance over ScanNet.
Finally, we estimate per-point normals using local PCA. This process aligns VGPC, converting them into geometrically consistent pre-training data.

\noindent{\textbf{Dataset statistics.}}
The final \data{} contains 49,219 scenes from VGPC, covering diverse lighting conditions, camera trajectories, and architectural styles across different countries. Compared to existing 3D datasets, \data{} captures a greater diversity of 3D structures, providing a scalable foundation for learning stable representations from unlabeled videos. Here, we denote the full version, which contains 49,219 VGPC including videos from existing datasets, as \data-49k (or \method{}-49k for the model trained on \data-49k). We also refer to the subset consisting of 15,921 VGPC built solely from our collected videos as \data-16k (or \method{}-16k for the model trained on \data-16k). We provide the dataset details in the supplementary materials.
\section{Experiments}
In this section, we evaluate the effectiveness of \method{} through comprehensive experiments. 
\begin{table*}[t]
\vspace{-.2em}
\centering

\subfloat[
\textbf{Z-axis up alignment}. Z-axis alignment improves LP by 5.3 points.
\label{tab:z-axis_alignment}
]{
\begin{minipage}[t]{0.29\linewidth}\vspace{0pt}
\begin{center}
\tablestyle{4pt}{1.05}
\begin{tabular}{x{48}x{32}x{32}}
case & LP & Full-FT \\
\shline
w/o align.  & 51.8 & \textbf{76.1} \\
w/ align.  & \baseline{\textbf{57.1}} & \baseline{75.1} \\
\end{tabular}
\end{center}
\end{minipage}
} \hspace{2em}%
\subfloat[
\textbf{Scale alignment}. RoomTours dataset improves performance by aligning scales.
\label{tab:scale_alignment}
]{
\begin{minipage}[t]{0.29\linewidth}\vspace{0pt}
\begin{center}
\tablestyle{4pt}{1.05}
\begin{tabular}{x{48}x{32}x{32}}
case & LP & Full-FT \\
\shline
w/o align.  & 55.5 & \textbf{75.3} \\
w/ align.  & \baseline{\textbf{57.1}} & \baseline{75.1} \\
\end{tabular}
\end{center}
\end{minipage}
}\hspace{2em}%
\subfloat[
\textbf{Self-distillation}. A self-distillation with noise regularization is more accurate.
\label{tab:self-distillation}
]{
\begin{minipage}[t]{0.29\linewidth}\vspace{0pt}
\begin{center}
\tablestyle{4pt}{1.05}
\begin{tabular}{x{64}x{32}x{32}}
case & LP & Full-FT \\
\shline
w/o regularizer  & 51.1 & 74.4 \\
w/ regularizer  & \baseline{\textbf{57.1}} & \baseline{\textbf{75.1}} \\
\end{tabular}
\end{center}
\end{minipage}
}
\\[-0.2em]
\vspace{.3em}

\subfloat[
\textbf{Regularizer loss}. Both $R_\text{Lap}$ and $R_\text{cons}$ work to improve the pre-training effect.
\label{tab:regularizer_loss}
]{
\begin{minipage}[t]{0.29\linewidth}\vspace{0pt}
\begin{center}
\tablestyle{4pt}{1.05}
\begin{tabular}{x{72}x{32}x{32}}
case & LP & Full-FT \\
\shline
w/o $R_\text{Lap}$  & 55.0 & 74.8 \\
w/o $R_\text{cons}$  & 55.3 & 75.0 \\
w/ $R_\text{Lap}$ \& $R_\text{cons}$  & \baseline{\textbf{57.1}} & \baseline{\textbf{75.1}} \\
\end{tabular}
\end{center}
\end{minipage}
}\hspace{2em}%
\subfloat[
\textbf{Data scaling}. RoomTours \#data scaling is linked to enhanced recognition performance.
\label{tab:data_scaling}
]{
\begin{minipage}[t]{0.29\linewidth}\vspace{0pt}
\begin{center}
\tablestyle{4pt}{1.05}
\begin{tabular}{x{48}x{32}x{32}}
\#scene & LP & Full-FT \\
\shline
1k  & \baseline{57.1} & \baseline{75.1} \\
16k  & 58.9 & 75.6\\
49k  & \textbf{65.9} & \textbf{77.7} \\
\end{tabular}
\end{center}
\end{minipage}
}\hspace{2em}%
\subfloat[
\textbf{Reconstruction model}. RoomTours construction using $\pi^3$ generally performs well.
\label{tab:reconstruct-model}
]{
\begin{minipage}[t]{0.3\linewidth}\vspace{0pt}
\begin{center}
\tablestyle{1pt}{1.05}
\begin{tabular}{x{70}x{32}x{32}}
reconst. model & LP & Full-FT \\
\shline
VGGT~\cite{wang2025vggt} & 49.3 & 75.1 \\
MapAnything~\cite{keetha2025mapanything} & 50.1 & \textbf{75.5} \\
$\pi^3$~\cite{wang2025pi} & \baseline{\textbf{57.1}} & \baseline{75.1} \\
\end{tabular}
\end{center}
\end{minipage}
}
\vspace{-.1em}
\caption{\textbf{\method ablation experiments} with PTv3 on ScanNet semantic segmentation benchmark. We report full fine-tuning (Full-FT) and linear probing (LP) performance (\%). Unless otherwise specified, the default settings use the \data generated by using $\pi^3$ and pretrain it using PTv3 (Base). Full-FT and LP are set to 100 epochs. Default settings for this ablation are marked in \colorbox{baselinecolor}{gray}.}
\label{tab:ablations}
\vspace{-10pt}
\end{table*}

\subsection{Implementation Details}
\label{method:implementation}
\noindent{\textbf{Pre-training configuration.}}
We pre-train the model using \method{} with a Point Transformer V3 (PTv3)~\cite{Wu_2024_CVPR} backbone. Each scene provides 9D input features consisting of coordinates, colors and normals. Training is conducted on eight NVIDIA H200 GPUs with a total batch size of 16. AdamW~\cite{loshchilov2017decoupled} is used as the optimizer with an initial learning rate of 0.001 and layer-wise decay of 0.9. The OneCycleLR~\cite{smith2018disciplined} learning rate scheduler is used, running for 145,600 iterations with a warm-up period followed by cosine annealing. Weight decay is linearly increased from 0.04 to 0.1 during pre-training. 
For the default RoomTours-1k / PTv3-Base setting, we train for 145,600 iterations with OneCycleLR. For larger-scale settings, we scale the number of iterations with dataset and model size (see Supp. Tab.~\ref{tab:lam3c_config}).

\noindent{\textbf{Evaluation protocol.}}
We evaluate the pre-training effectiveness of \method{} on semantic and instance segmentation tasks in indoor scenes. We use four widely used indoor scene datasets for evaluation: ScanNet~\cite{dai2017scannet}, ScanNet++~\cite{yeshwanth2023scannet++}, ScanNet200~\cite{rozenberszki2022language}, and S3DIS~\cite{Armeni_2016_CVPR}. Following standard evaluation protocols, we report results under two settings: (i) Full fine-tuning, in which all pre-trained model parameters are updated on the fine-tuning dataset, and (ii) Linear probing, in which the backbone parameters are frozen and only a final linear layer is trained on the fine-tuning dataset. The details of the hyperparameters for each dataset are described in the supplementary material.

\noindent{\textbf{Comparison baseline.}}
We use Sonata as a baseline in our experiments. We noticed that previous 3D-SSL techniques could include, during pre-training, the validation and test splits of datasets also used for downstream evaluation. This is the case for Sonata, in particular.\footnote{\url{https://github.com/facebookresearch/sonata/issues/35}} While no labels of the validation and test splits are accessed, including these splits during pre-training nevertheless improves performance, as shown in the supplementary material. As our method uses no real point clouds, let alone validation or test data, we remove these splits when pre-training Sonata baselines to ensure fair comparison.

\subsection{Main results}
\label{sec:main_results}
This section presents comparative experiments on semantic and instance segmentation tasks on indoor scene datasets, comparing our method against state-of-the-art approaches.

\noindent{\textbf{Semantic segmentation.}}
Here, we evaluate the effect of pre-training on downstream semantic segmentation tasks using mIoU as the evaluation metric. The results are shown in Table~\ref{tab:indoor-semseg}. First, compared to training PTv3 from scratch, \method{} improves performance across all datasets under both linear probing and full fine-tuning settings. This shows that effective representations can be acquired even through pre-training using only VGPC. 

This suggests that the VGPC provide complementary information rather than simply replacing real data. Based on these results, \method{} can be considered an effective pre-training method for semantic segmentation tasks in indoor scenes.

\noindent{\textbf{Instance segmentation.}}
We also evaluate the pre-training effect on instance segmentation tasks.
AP is used as the evaluation metric, and the results are shown in Table~\ref{tab:indoor-insseg}. First, compared to training PTv3 from scratch, \method{} consistently improves AP across all datasets under both linear probing and full fine-tuning settings. This demonstrates that representations enabling individual object recognition can be acquired even through pre-training without real 3D scans. 
This suggests that VGPC provide complementary information regarding the geometric distinctiveness required for instance segmentation, rather than merely substituting real data. Therefore, \method{} can be considered an effective pre-training method for instance segmentation in indoor scenes.

\subsection{Ablation studies}
\label{sec:abalation_study}
This section explores the key components of the proposed regularized loss and the \data dataset, aiming to identify effective elements of \method{}.

\noindent{\textbf{Effect of z-axis alignment (see Table~\ref{tab:z-axis_alignment}).}} 
This experiment aims to investigate the effectiveness of the z-axis alignment for \data. Table~\ref{tab:z-axis_alignment} shows that  \data with the z-axis alignment improves segmentation performance. For linear probing in particular, \data with z-axis alignment performs better by +5.3 points. Thus, z-axis alignment is a key factor, especially when using frozen pre-trained model parameters produced by \method{}.

\noindent{\textbf{Effect of scale alignment (see Table~\ref{tab:scale_alignment}).}}
This experiment aims to evaluate the effectiveness of scale alignment on \data. We adjust each scene in \data to align roughly with the scale distribution of ScanNet. Table~\ref{tab:scale_alignment} confirms that \data with scale alignment achieves a performance improvement of +1.6 points, particularly in linear probing. This suggests that data scale impacts performance in LP, given that parameters other than the final layer of the pre-trained model are frozen. Based on these results, our scale alignment for \data contributes to the effectiveness of pre-training.

\noindent{\textbf{Effect of \method{} components (see Table~\ref{tab:self-distillation} and Table~\ref{tab:regularizer_loss}).}} 
This experiment verifies the effectiveness of noise-regularized loss of \method{}. First, we compare the results of using only the clustering loss with those of adding our proposed noise-regularized loss. Table~\ref{tab:self-distillation} confirms that noise-regularized loss achieves an improvement in segmentation performance compared to clustering loss alone, for both LP and Full-FT. This suggests that noise-regularized loss is effective for \data composed of VGPC containing noise and missing regions, and helps stabilize representation learning.
Furthermore, we investigate the combined effects of the Laplacian smoothing loss and the noise consistency loss in \method{}. Table~\ref{tab:regularizer_loss} shows that noise-regularized loss using both loss functions achieves higher pre-training effectiveness than using either Laplacian smoothing loss or noise consistency loss alone. This result indicates that the Laplacian smoothing loss and the noise consistency loss play complementary roles in promoting stable learning on noisy point clouds.

\noindent{\textbf{Effect of data scaling (see Table~\ref{tab:data_scaling}).}}
\data enables scalable VGPC generation from videos with limited manual intervention, making it easier to scale than real 3D scans, provided that video and computational resources are available. This experiment investigates the effect of pre-training on the number of VGPC scenes in \data. Compared to 1k scenes, increasing the number of scenes to 16k and 49k tends to improve performance. These results demonstrate that \method{}, despite not using real 3D scans, holds potential for scaling with VGPC.

\noindent{\textbf{Effect of reconstruction model (see Table~\ref{tab:reconstruct-model}).}}
We assess the impact of feed-forward reconstruction models on VGPC. This experiment compares VGGT, MapAnything, and $\pi^3$. Table~\ref{tab:reconstruct-model} shows that $\pi^3$ exhibits greater pre-training effectiveness than VGGT and MapAnything. This is because $\pi^3$ can process longer sequences, enabling it to reconstruct high-quality VGPC compared with VGGT and MapAnything.

\subsection{Additional Experiments}
\label{sec:additional_exp}
We evaluate the pre-training performance of \method{} under limited resources. Furthermore, based on qualitative results, we analyze the zero-shot segmentation performance.

\noindent{\textbf{Impact of limited resources.}} 
In this experiment, we evaluate the effect of \method{} under limited resources for downstream tasks. On ScanNet semantic segmentation, we restrict the number of training scenes to \{1\%, 5\%, 10\%, 20\%\}, and separately limit the number of annotated point clouds to \{20, 50, 100, 200\}. As shown in Table~\ref{tab:data-efficiency}, our method provides clear gains even when both the number of training samples and the amount of supervision are heavily constrained. LAM3C remains competitive under limited supervision and consistently outperforms the ScanNet only Sonata baseline, while approaching or exceeding stronger Sonata variants in several settings.

\begin{table}[t]
\centering
\caption{Comparison of semantic segmentation performance on limited training samples and annotation in ScanNet. }
\vspace{-6pt}
\label{tab:data-efficiency}
\scalebox{0.77}{
\begin{tabular}{lrrrrrrrrrrr}
\toprule
Data Efficiency &\multicolumn{4}{c}{Limited Scenes (Pct.)} &\multicolumn{4}{c}{Limited Annotation (Pts.)} \\
\cmidrule(lr){1-1} \cmidrule(lr){2-5} \cmidrule(lr){6-9}
Methods &1\% &5\% &10\% &20\% &20 &50 &100 &200 \\
\midrule
PTv3 & 23.6 & 47.2 & 59.7 & 67.2 & 62.2 & 68.1 & 70.6 & 71.8 \\
\midrule
\rowcolor[gray]{0.9} PPT~\cite{wu2024towards}~(sup.) & \textcolor{darkgray}{31.1} & \textcolor{darkgray}{52.6} &
\textcolor{darkgray}{63.3}  & \textcolor{darkgray}{68.2} & 
\textcolor{darkgray}{62.4}  & \textcolor{darkgray}{69.1} & \textcolor{darkgray}{74.3} & \textcolor{darkgray}{75.5}   \\
\rowcolor[gray]{0.9} Sonata~\cite{wu2025sonata}~ & \textcolor{darkgray}{45.3} & \textcolor{darkgray}{65.7} &
\textcolor{darkgray}{72.4}  & \textcolor{darkgray}{72.8} & 
\textcolor{darkgray}{70.5}  & \textcolor{darkgray}{73.6} & \textcolor{darkgray}{76.0} & \textcolor{darkgray}{77.0}   \\
\midrule
Sonata (all real) & \textbf{43.1} & \textbf{63.1} & 70.0 & \textbf{71.7} & 69.9 & 73.9 & 75.3 & 76.4  \\
Sonata (ScanNet) & 36.0 & 57.7 & 66.7 & 68.2 & 66.3 & 70.2 & 71.4 & 73.0  \\

\rowcolor[HTML]{DEF3E7} \method{}-16k  & \cellcolor[HTML]{def3e6}36.8 & \cellcolor[HTML]{def3e6}56.5 &\cellcolor[HTML]{def3e6}65.8 & \cellcolor[HTML]{def3e6}68.6  & \cellcolor[HTML]{def3e6}66.0 & \cellcolor[HTML]{def3e6}70.2 & \cellcolor[HTML]{def3e6}73.3 & \cellcolor[HTML]{def3e6}73.7 \\

\rowcolor[HTML]{DEF3E7} \method{}-49k  & \cellcolor[HTML]{def3e6}40.1 & \cellcolor[HTML]{def3e6} 59.2 &\cellcolor[HTML]{def3e6} 68.7 & \cellcolor[HTML]{def3e6}71.6  & \cellcolor[HTML]{def3e6}68.8 & \cellcolor[HTML]{def3e6}73.0 & \cellcolor[HTML]{def3e6}75.1 & \cellcolor[HTML]{def3e6}76.2 \\

\rowcolor[HTML]{DEF3E7} \method{}-49k*  & \cellcolor[HTML]{def3e6}40.5 & \cellcolor[HTML]{def3e6} 60.1 &\cellcolor[HTML]{def3e6}\textbf{71.9} & \cellcolor[HTML]{def3e6} 71.6  & \cellcolor[HTML]{def3e6}\textbf{70.4} & \cellcolor[HTML]{def3e6}\textbf{74.2} & \cellcolor[HTML]{def3e6}\textbf{76.2} & \cellcolor[HTML]{def3e6}\textbf{77.1} \\

\bottomrule
\multicolumn{9}{l}{* uses PTv3 (Large) and 437k pre-training steps. }
\end{tabular}}
\end{table}

\begin{table}[t]
\centering
\caption{Synergy between \data and \method}
\vspace{-10pt}
\label{tab:synergy}
\centering
\renewcommand{\arraystretch}{0.95}
\begin{tabular}{lcc}
\toprule
 & Sonata & {\text{\method}} \\
\midrule
ScanNet (clean) & 67.1 & 66.8 \\
\data (noisy) & 51.1 & 57.1 \\
\bottomrule
\end{tabular}
\vspace{-10pt}
\end{table}

\noindent{\textbf{Disentangling the contributions of \method and \data.}}
When pre-trained on ScanNet, the two self-supervised methods, Sonata and \method, perform comparably, indicating that our proposed self-supervised method does not provide inherent gains on curated 3D scans, as shown in Table~\ref{tab:synergy}. 
In contrast, on \data-1k, \method substantially outperforms Sonata, indicating that the improvement arises from the learning design rather than the dataset alone. 
Taken together, \data provides scalable scan-free pre-training data, while \method is essential for learning effectively from noisy reconstructed point clouds.

\begin{figure}[t]
  \centering
  \includegraphics[width=\linewidth]{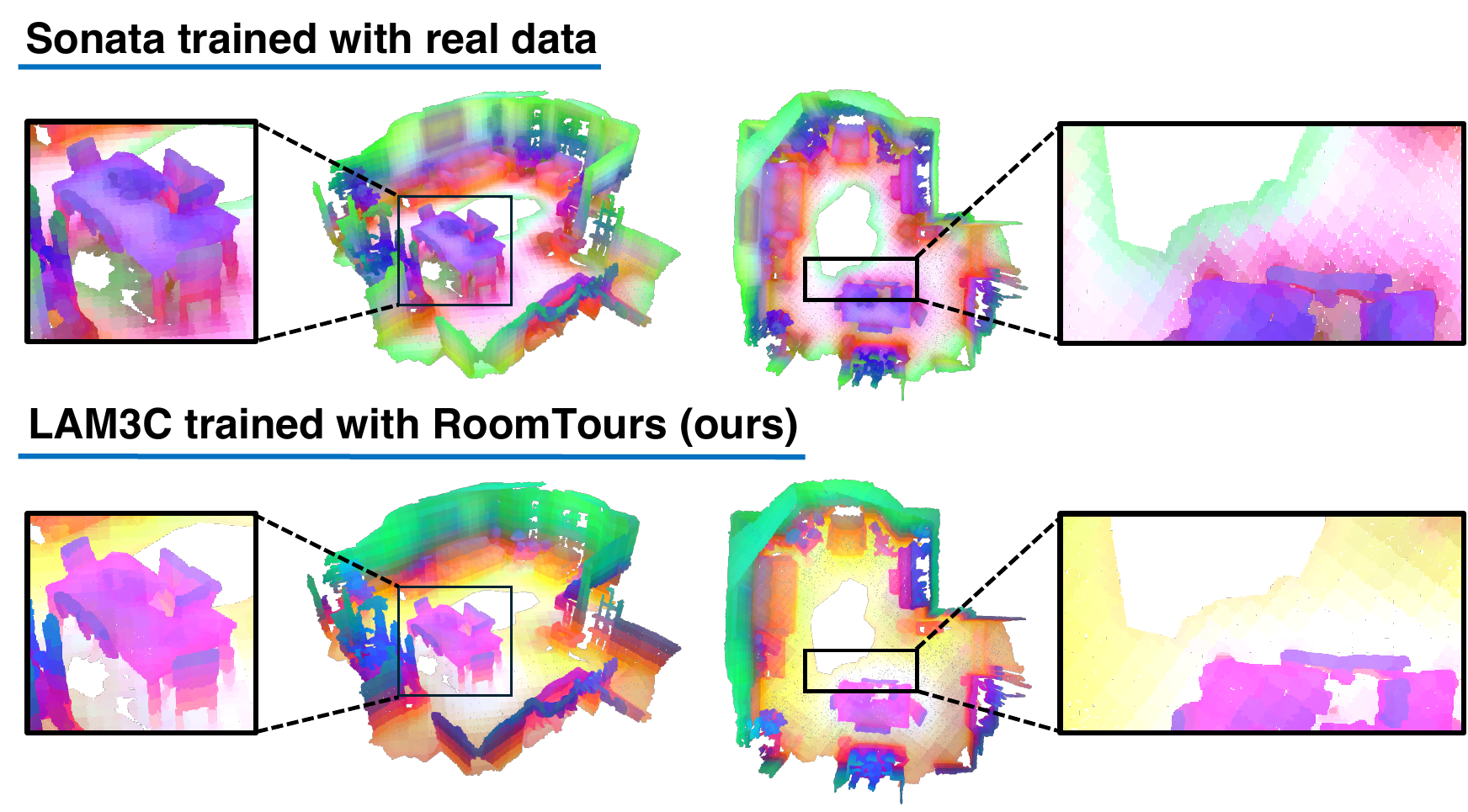}
  \caption{\textbf{Zero-shot qualitative evaluation using PCA visualizations.} LAM3C learns semantic representations in the real world without pre-training on real 3D scans.}
  \label{fig:qualitative_results}
  \vspace{-10pt}
\end{figure}

\noindent{\textbf{Qualitative results.}} 
This experiment qualitatively evaluates zero-shot recognition by using PCA. As shown in Fig.~\ref{fig:qualitative_results}, \method{} produces clear segmentation of local structures such as desks even in a zero-shot setting. This result indicates that VGPC preserve enough coarse geometric cues to learn representations for 3D-SSL directly from video. At the same time,  \method{} exhibits less coherent global structure compared to Sonata trained with real 3D scans. For example, floor edges show lower contrast, and the floor boundaries appear more blurred in the PCA space. This behavior is expected because $\pi^3$ reconstructions provide approximate 3D representations with variations in coordinate frames and scale, which makes it more difficult for the model to learn globally consistent scene geometry than when the model is pre-trained on accurate 3D scans.

\section{Conclusion}
We have presented \method{}, a self-supervised framework that learns 3D representations from unlabeled videos. By leveraging VGPC reconstructed with a feed-forward reconstruction model and introducing noise-regularized objectives, namely local smoothness and noise consistency, \method{} achieves comparable or better pre-training performance without using any real 3D scans. We also constructed \data{}, a dataset of 49k VGPC derived from unlabeled room-tour videos, and showed that \method{} matches or surpasses prior pre-training methods such as Sonata, even under limited downstream data conditions. These results demonstrate that unlabeled videos can serve as an effective and scalable source of supervision for 3D-SSL.
Our contribution is this scan-free and scalable 3D-SSL setting based on video-only data collection, rather than the input format itself. Unlike prior 3D-SSL methods, we generate pre-training data from unlabeled videos at scale. 

\section*{Acknowledgment}
This work was supported by the AIST policy-based budget project “R\&D on Generative AI Foundation Models for the Physical Domain.” It was also supported by the JSPS Overseas Research Fellowship. In addition, this work was supported by the Japan Science and Technology Agency (JST) through the Adopting Sustainable Partnerships for Innovative Research Ecosystem (ASPIRE) program under Grant Number JPMJAP2518. We used ABCI 3.0, provided by AIST and AIST Solutions, with support from the “ABCI 3.0 Development Acceleration Use” program.

{
    \small
    \bibliographystyle{ieeenat_fullname}
    \bibliography{main}

@String(IJCV = {Int. J. Comput. Vis.})

@String(CVPR= {IEEE Conf. Comput. Vis. Pattern Recog.})

@String(ICCV= {Int. Conf. Comput. Vis.})

@String(ECCV= {Eur. Conf. Comput. Vis.})

@String(TOG= {ACM Trans. Graph.})

@String(ICLR = {Int. Conf. Learn. Represent.})

@String(IJCV  = {IJCV})

@String(CVPR  = {CVPR})

@String(ICCV  = {ICCV})

@String(ECCV  = {ECCV})

@String(TOG   = {ACM TOG})

@String(ICLR  = {ICLR})

@inproceedings{chen2021shape,
    title={Shape self-correction for unsupervised point cloud understanding},
    author={Chen, Ye and Liu, Jinxian and Ni, Bingbing and Wang, Hang and Yang, Jiancheng and Liu, Ning and Li, Teng and Tian, Qi},
    booktitle={ICCV},
    year={2021}
}

@inproceedings{poursaeed2020self,
    title={Self-supervised learning of point clouds via orientation estimation},
    author={Poursaeed, Omid and Jiang, Tianxing and Qiao, Han and Xu, Nayun and Kim, Vladimir G},
    booktitle={3DV},
    year={2020}
}

@inproceedings{sauder2019self,
    title={Self-supervised deep learning on point clouds by reconstructing space},
    author={Sauder, Jonathan and Sievers, Bjarne},
    booktitle={NeurIPS},
    year={2019}
}

@inproceedings{depthcontrast,
    author = {Zhang, Zaiwei and Girdhar, Rohit and Joulin, Armand and Misra, Ishan},
    title = {Self-Supervised Pretraining of {3D} Features on Any Point-Cloud},
    booktitle={ICCV},
    year = {2021},
}

@article{segcontrast,
    author = {Lucas Nunes and Rodrigo Marcuzzi and Xieyuanli Chen and Jens Behley and Cyrill Stachniss},
    title = {{SegContrast}: {3D} Point Cloud Feature Representation Learning through Self-supervised Segment Discrimination},
    journal = {RA-L},
    year={2022},
}

@inproceedings{tarl,
    author = {Lucas Nunes and Louis Wiesmann and Rodrigo Marcuzzi and Xieyuanli Chen and Jens Behley and Cyrill Stachniss},
    title = {Temporal Consistent {3D} LiDAR Representation Learning for Semantic Perception in Autonomous Driving},
    booktitle={CVPR},
    year = {2023}
}

@inproceedings{proposalcontrast,
    title={{ProposalContrast}: Unsupervised pre-training for lidar-based {3D} object detection},
    author={Yin, Junbo and Zhou, Dingfu and Zhang, Liangjun and Fang, Jin and Xu, Cheng-Zhong and Shen, Jianbing and Wang, Wenguan},
    booktitle={ECCV},
    year={2022}
}

@inproceedings{liu2023fac,
    title={{FAC}: {3D} representation learning via foreground aware feature contrast},
    author={Liu, Kangcheng and Xiao, Aoran and Zhang, Xiaoqin and Lu, Shijian and Shao, Ling},
    booktitle={CVPR},
    year = 2023,
}

@inproceedings{strl,
    title={Spatio-temporal self-supervised representation learning for {3D} point clouds},
    author={Huang, Siyuan and Xie, Yichen and Zhu, Song-Chun and Zhu, Yixin},
    booktitle={ICCV},
    year={2021}
}

@inproceedings{bevcontrast,
  author    = {Corentin Sautier and Gilles Puy and Alexandre Boulch and Renaud Marlet and Vincent Lepetit},
  title     = {{BEVContrast}: Self-Supervision in BEV Space for Automotive Lidar Point Clouds},
  booktitle={3DV},
  year      = 2024
}

@inproceedings{groupcontrast,
    author    = {Wang, Chengyao and Jiang, Li and Wu, Xiaoyang and Tian, Zhuotao and Peng, Bohao and Zhao, Hengshuang and Jia, Jiaya},
    title     = {GroupContrast: Semantic-aware Self-supervised Representation Learning for 3D Understanding},
    booktitle = {CVPR},
    year      = {2024},
}

@InProceedings{Hou_2021_CVPR,
    author    = {Hou, Ji and Graham, Benjamin and Niessner, Matthias and Xie, Saining},
    title     = {Exploring Data-Efficient 3D Scene Understanding With Contrastive Scene Contexts},
    booktitle={CVPR},
    year      = {2021},
}

@article{voxelmae1,
    title={{Voxel-MAE}: Masked autoencoders for pre-training large-scale point clouds},
    author={Min, Chen and Zhao, Dawei and Xiao, Liang and Nie, Yiming and Dai, Bin},
    journal={arXiv preprint arXiv:2206.09900},
    year={2022}
}

@inproceedings{voxelmae2,
    title={Masked Autoencoder for Self-Supervised Pre-Training on Lidar Point Clouds},
    author={Hess, Georg and Jaxing, Johan and Svensson, Elias and Hagerman, David and Petersson, Christoffer and Svensson, Lennart},
    booktitle={WACV},
    year={2023}
}

@inproceedings{pointbert,
    title={{Point-BERT}: Pre-training {3D} point cloud transformers with masked point modeling},
    author={Yu, Xumin and Tang, Lulu and Rao, Yongming and Huang, Tiejun and Zhou, Jie and Lu, Jiwen},
    booktitle={CVPR},
    year={2022}
}

@inproceedings{pointm2ae,
    title={{Point-M2AE}: multi-scale masked autoencoders for hierarchical point cloud pre-training},
    author={Zhang, Renrui and Guo, Ziyu and Gao, Peng and Fang, Rongyao and Zhao, Bin and Wang, Dong and Qiao, Yu and Li, Hongsheng},
    booktitle={NeurIPS},
    year={2022}
}

@inproceedings{pointmae,
    title={Masked autoencoders for point cloud self-supervised learning},
    author={Pang, Yatian and Wang, Wenxiao and Tay, Francis EH and Liu, Wei and Tian, Yonghong and Yuan, Li},
    booktitle={ECCV},
    year={2022}
}

@inproceedings{also,
    title = {{ALSO}: Automotive Lidar Self-supervision by Occupancy estimation},
    author = {Alexandre Boulch and Corentin Sautier and Björn Michele and Gilles Puy and Renaud Marlet},
    booktitle={CVPR},
    year = 2023,
}

@inproceedings{caron2021emerging,
  title={Emerging properties in self-supervised vision transformers},
  author={Caron, Mathilde and Touvron, Hugo and Misra, Ishan and J{\'e}gou, Herv{\'e} and Mairal, Julien and Bojanowski, Piotr and Joulin, Armand},
  booktitle={ICCV},
  year={2021}
}

@article{oquab2023dinov2,
  title={Dinov2: Learning robust visual features without supervision},
  author={Oquab, Maxime and Darcet, Timoth{\'e}e and Moutakanni, Th{\'e}o and Vo, Huy and Szafraniec, Marc and Khalidov, Vasil and Fernandez, Pierre and Haziza, Daniel and Massa, Francisco and El-Nouby, Alaaeldin and others},
  journal={TMLR},
  year={2023}
}

@article{simeoni2025dinov3,
  title={Dinov3},
  author={Sim{\'e}oni, Oriane and Vo, Huy V. and Seitzer, Maximilian and Baldassarre, Federico and Oquab, Maxime and Jose, Cijo and Khalidov, Vasil and Szafraniec, Marc and Yi, Seungeun and Ramamonjisoa, Micha{\"e}l and Massa, Francisco and Haziza, Daniel and Wehrstedt, Luca and Wang, Jianyuan and Darcet, Timoth{\'e}e and Moutakanni, Th{\'e}o and Sentana, Leonel and Roberts, Claire and Vedaldi, Andrea and Tolan, Jamie and Brandt, John and Couprie, Camille and Mairal, Julien and J{\'e}gou, Herv{\'e} and Labatut, Patrick and Bojanowski, Piotr},
  journal={arXiv preprint arXiv:2508.10104},
  year={2025}
}

@InProceedings{Kirillov_2023_ICCV,
    author    = {Kirillov, Alexander and Mintun, Eric and Ravi, Nikhila and Mao, Hanzi and Rolland, Chloe and Gustafson, Laura and Xiao, Tete and Whitehead, Spencer and Berg, Alexander C. and Lo, Wan-Yen and Dollar, Piotr and Girshick, Ross},
    title     = {Segment Anything},
    booktitle = {ICCV},
    year      = {2023},
}

@inproceedings{wu2025sonata,
  title={Sonata: Self-supervised learning of reliable point representations},
  author={Wu, Xiaoyang and DeTone, Daniel and Frost, Duncan and Shen, Tianwei and Xie, Chris and Yang, Nan and Engel, Jakob and Newcombe, Richard and Zhao, Hengshuang and Straub, Julian},
  booktitle={CVPR},
  year={2025}
}

@inproceedings{wang2025vggt,
  title={Vggt: Visual geometry grounded transformer},
  author={Wang, Jianyuan and Chen, Minghao and Karaev, Nikita and Vedaldi, Andrea and Rupprecht, Christian and Novotny, David},
  booktitle={CVPR},
  year={2025}
}

@inproceedings{wu2024towards,
  title={Towards large-scale 3D representation learning with multi-dataset point prompt training},
  author={Wu, Xiaoyang and Tian, Zhuotao and Wen, Xin and Peng, Bohao and Liu, Xihui and Yu, Kaicheng and Zhao, Hengshuang},
  booktitle={CVPR},
  year={2024}
}

@inproceedings{wu2023masked,
  title={Masked scene contrast: A scalable framework for unsupervised 3D representation learning},
  author={Wu, Xiaoyang and Wen, Xin and Liu, Xihui and Zhao, Hengshuang},
  booktitle={CVPR},
  year={2023}
}

@inproceedings{xie2020pointcontrast,
  title={Pointcontrast: Unsupervised pre-training for 3D point cloud understanding},
  author={Xie, Saining and Gu, Jiatao and Guo, Demi and Qi, Charles R and Guibas, Leonidas and Litany, Or},
  booktitle={ECCV},
  year={2020},
}

@article{ullman1979interpretation,
  title={The interpretation of structure from motion},
  author={Ullman, Shimon},
  journal={Proceedings of the Royal Society of London. Series B. Biological Sciences},
  year={1979},
  }

@inproceedings{yang2025thinking,
  title={Thinking in space: How multimodal large language models see, remember, and recall spaces},
  author={Yang, Jihan and Yang, Shusheng and Gupta, Anjali W and Han, Rilyn and Fei-Fei, Li and Xie, Saining},
  booktitle={CVPR},
  year={2025}
}

@InProceedings{Wang_2024_CVPR,
    author    = {Wang, Shuzhe and Leroy, Vincent and Cabon, Yohann and Chidlovskii, Boris and Revaud, Jerome},
    title     = {DUSt3R: Geometric 3D Vision Made Easy},
    booktitle = {CVPR},
    year      = {2024},
}

@inproceedings{dai2017scannet,
    title={ScanNet: Richly-annotated 3D Reconstructions of Indoor Scenes},
    author={Dai, Angela and Chang, Angel X. and Savva, Manolis and Halber, Maciej and Funkhouser, Thomas and Nie{\ss}ner, Matthias},
    booktitle = {CVPR},
    year = {2017}
}

@article{wang2025pi,
  title={$\pi^3$: Permutation-Equivariant Visual Geometry Learning},
  author={Wang, Yifan and Zhou, Jianjun and Zhu, Haoyi and Chang, Wenzheng and Zhou, Yang and Li, Zizun and Chen, Junyi and Pang, Jiangmiao and Shen, Chunhua and He, Tong},
  journal={arXiv preprint arXiv:2507.13347},
  year={2025}
}

@misc{wu2011visualsfm,
  title={VisualSFM: A visual structure from motion system},
  author={Wu, Changchang and others},
  year={2011},
  publisher={St. Louis, MO, USA}
}

@article{agarwal2011building,
  title={Building rome in a day},
  author={Agarwal, Sameer and Furukawa, Yasutaka and Snavely, Noah and Simon, Ian and Curless, Brian and Seitz, Steven M and Szeliski, Richard},
  journal={CACM},
  year={2011},
}

@article{tomasi1992shape,
  title={Shape and motion from image streams under orthography: a factorization method},
  author={Tomasi, Carlo and Kanade, Takeo},
  journal={IJCV},
  year={1992},
}

@article{okutomi1993multiple,
  title={A multiple-baseline stereo},
  author={Okutomi, Masatoshi and Kanade, Takeo},
  journal={TPAMI},
  year={1993},
}

@article{keetha2025mapanything,
  title={MapAnything: Universal feed-forward metric 3D reconstruction},
  author={Keetha, Nikhil and M{\"u}ller, Norman and Sch{\"o}nberger, Johannes and Porzi, Lorenzo and Zhang, Yuchen and Fischer, Tobias and Knapitsch, Arno and Zauss, Duncan and Weber, Ethan and Antunes, Nelson and Jonathon, Luiten and Manuel, Lopez-Antequera and Samuel, Rota Bulò and Christian, Richardt and Deva, Ramanan and Sebastian, Scherer and Peter, Kontschieder},
  journal={arXiv preprint arXiv:2509.13414},
  year={2025}
}

@article{furukawa2009accurate,
  title={Accurate, dense, and robust multiview stereopsis},
  author={Furukawa, Yasutaka and Ponce, Jean},
  journal={TPAMI},
  year={2009},
}

@article{seitz1999photorealistic,
  title={Photorealistic scene reconstruction by voxel coloring},
  author={Seitz, Steven M and Dyer, Charles R},
  journal={IJCV},
  year={1999},
}

@inproceedings{triggs1999bundle,
  title={Bundle adjustment—a modern synthesis},
  author={Triggs, Bill and McLauchlan, Philip F and Hartley, Richard I and Fitzgibbon, Andrew W},
  booktitle={IWVA},
  year={1999},
  }

@inproceedings{cheeseman1987stochastic,
  title={A stochastic map for uncertain spatial relationships},
  author={Cheeseman, Peter and Smith, Robert and Self, Michael},
  booktitle={ISRR},
  year={1987},
}

@article{davison2007monoslam,
  title={MonoSLAM: Real-time single camera SLAM},
  author={Davison, Andrew J and Reid, Ian D and Molton, Nicholas D and Stasse, Olivier},
  journal={TPAMI},
  year={2007},
}

@inproceedings{radford2021learning,
  title={Learning transferable visual models from natural language supervision},
  author={Radford, Alec and Kim, Jong Wook and Hallacy, Chris and Ramesh, Aditya and Goh, Gabriel and Agarwal, Sandhini and Sastry, Girish and Askell, Amanda and Mishkin, Pamela and Clark, Jack and others},
  booktitle={ICML},
  year={2021},
}

@inproceedings{rozenberszki2022language,
  title={Language-grounded indoor 3D semantic segmentation in the wild},
  author={Rozenberszki, David and Litany, Or and Dai, Angela},
  booktitle={ECCV},
  year={2022},
  }

@inproceedings{yeshwanth2023scannet++,
  title={Scannet++: A high-fidelity dataset of 3D indoor scenes},
  author={Yeshwanth, Chandan and Liu, Yueh-Cheng and Nie{\ss}ner, Matthias and Dai, Angela},
  booktitle={ICCV},
  year={2023}
}

@article{polyak1992acceleration,
  title={Acceleration of stochastic approximation by averaging},
  author={Polyak, Boris T and Juditsky, Anatoli B},
  journal={SIAM journal on control and optimization},
  year={1992},
}

@inproceedings{cuturi2013sinkhorn,
  title={Sinkhorn distances: Lightspeed computation of optimal transport},
  author={Cuturi, Marco},
  booktitle={NeurIPS},
  year={2013}
}

@inproceedings{loshchilov2017decoupled,
  title={Decoupled weight decay regularization},
  author={Loshchilov, Ilya and Hutter, Frank},
  booktitle={ICLR},
  year={2019}
}

@InProceedings{Armeni_2016_CVPR,
author = {Armeni, Iro and Sener, Ozan and Amir Roshan Zamir and Jiang, Helen and Brilakis, Ioannis and Fischer, Martin and Savarese, Silvio},
title = {3D Semantic Parsing of Large-Scale Indoor Spaces},
booktitle = {CVPR},
year = {2016}
}

@article{smith2018disciplined,
  title={A disciplined approach to neural network hyper-parameters: Part 1--learning rate, batch size, momentum, and weight decay},
  author={Smith, Leslie N},
  journal={arXiv preprint arXiv:1803.09820},
  year={2018}
}

@InProceedings{Wu_2024_CVPR,
    author    = {Wu, Xiaoyang and Jiang, Li and Wang, Peng-Shuai and Liu, Zhijian and Liu, Xihui and Qiao, Yu and Ouyang, Wanli and He, Tong and Zhao, Hengshuang},
    title     = {Point Transformer V3: Simpler Faster Stronger},
    booktitle = {CVPR},
    year      = {2024},
}

@inproceedings{baruch2021arkitscenes,
  title={Arkitscenes: A diverse real-world dataset for 3D indoor scene understanding using mobile rgb-D data},
  author={Baruch, Gilad and Chen, Zhuoyuan and Dehghan, Afshin and Feigin, Yuri and Fu, Peter and Gebauer, Thomas and Kurz, Daniel and Dimry, Tal and Joffe, Brandon and Schwartz, Arik and Elad Shulman},
  booktitle={NeurIPS},
  year={2021}
}

@InProceedings{Cabon_2025_CVPR,
    author    = {Cabon, Yohann and Stoffl, Lucas and Antsfeld, Leonid and Csurka, Gabriela and Chidlovskii, Boris and Revaud, Jerome and Leroy, Vincent},
    title     = {MUSt3R: Multi-view Network for Stereo 3D Reconstruction},
    booktitle = {CVPR},
    year      = {2025},
}

@inproceedings{asano2019self,
  title={Self-labelling via simultaneous clustering and representation learning},
  author={Asano, Yuki Markus and Rupprecht, Christian and Vedaldi, Andrea},
  booktitle={ICLR},
  year={2020}
}

@inproceedings{caron2020unsupervised,
  title={Unsupervised learning of visual features by contrasting cluster assignments},
  author={Caron, Mathilde and Misra, Ishan and Mairal, Julien and Goyal, Priya and Bojanowski, Piotr and Joulin, Armand},
  booktitle={NeurIPS},
  year={2020}
}

@article{knight2008sinkhorn,
  title={The Sinkhorn--Knopp algorithm: convergence and applications},
  author={Knight, Philip A},
  journal={SIAM Journal on Matrix Analysis and Applications},
  year={2008},
}

@article{zhou2018stereo,
  title={Stereo magnification: learning view synthesis using multiplane images},
  author={Zhou, Tinghui and Tucker, Richard and Flynn, John and Fyffe, Graham and Snavely, Noah},
  journal={TOG},
  year={2018}
}

@inproceedings{chang2020semantic,
  title={Semantic visual navigation by watching youtube videos},
  author={Chang, Matthew and Gupta, Arjun and Gupta, Saurabh},
  booktitle={NeurIPS},
  year={2020}
}

@InProceedings{Celen_2025_ICCV,
    author    = {\c{C}elen, Ata and Pollefeys, Marc and Barath, Daniel and Armeni, Iro},
    title     = {HouseTour: A Virtual Real Estate A(I)gent},
    booktitle = {ICCV},
    year      = {2025},
}

@misc{pointcept2023,
  author       = {Pointcept Contributors},
  title        = {Pointcept: A Codebase for Point Cloud Perception Research},
  howpublished = {\url{https://github.com/Pointcept/Pointcept}},
  year         = {2023},
  note         = {Accessed: 2025-11-19}
}

@article{wu2022point,
  title={Point transformer v2: Grouped vector attention and partition-based pooling},
  author={Wu, Xiaoyang and Lao, Yixing and Jiang, Li and Liu, Xihui and Zhao, Hengshuang},
  journal={NeurIPS},
  year={2022}
}

@inproceedings{zhang2025concerto,
  title={Concerto: Joint 2D-3D Self-Supervised Learning Emerges Spatial Representations},
  author={Zhang, Yujia and Wu, Xiaoyang and Lao, Yixing and Wang, Chengyao and Tian, Zhuotao and Wang, Naiyan and Zhao, Hengshuang},
  booktitle={NeurIPS},
  year={2025}
}
}

\appendix

\twocolumn[{%
\renewcommand\twocolumn[1][]{#1}%

\begin{center}
    \centering
    \captionsetup{type=figure}
    \includegraphics[width=\textwidth]{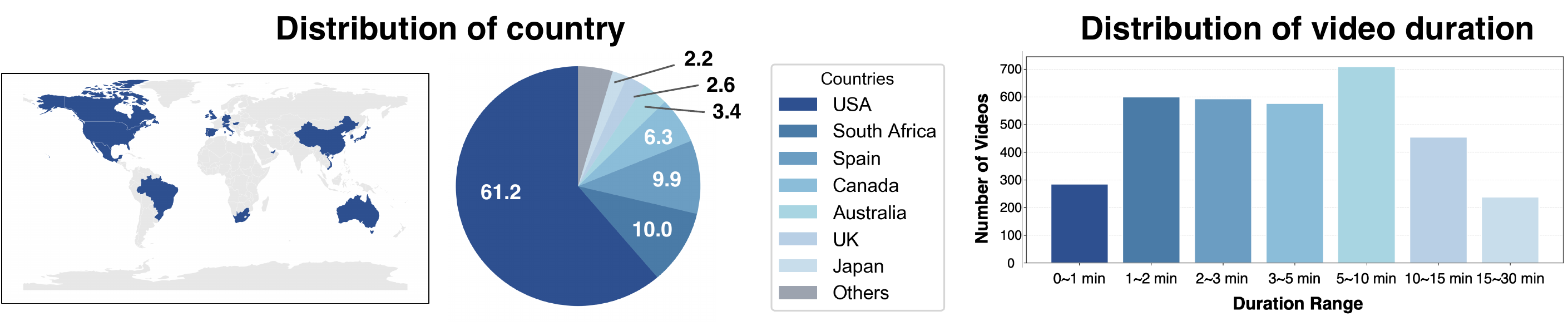}
    \caption{\textbf{Statistics of the \data video collection.} Our \data consists of 3,462 indoor walkthrough videos collected from 19 countries. (Left) Geographic distribution of the source countries and their proportions.  (Right) Distribution of video durations, ranging from short clips to long walkthroughs, with a median length of 3.63 minutes.}
    \label{fig:collect_video}

    \includegraphics[width=\textwidth]{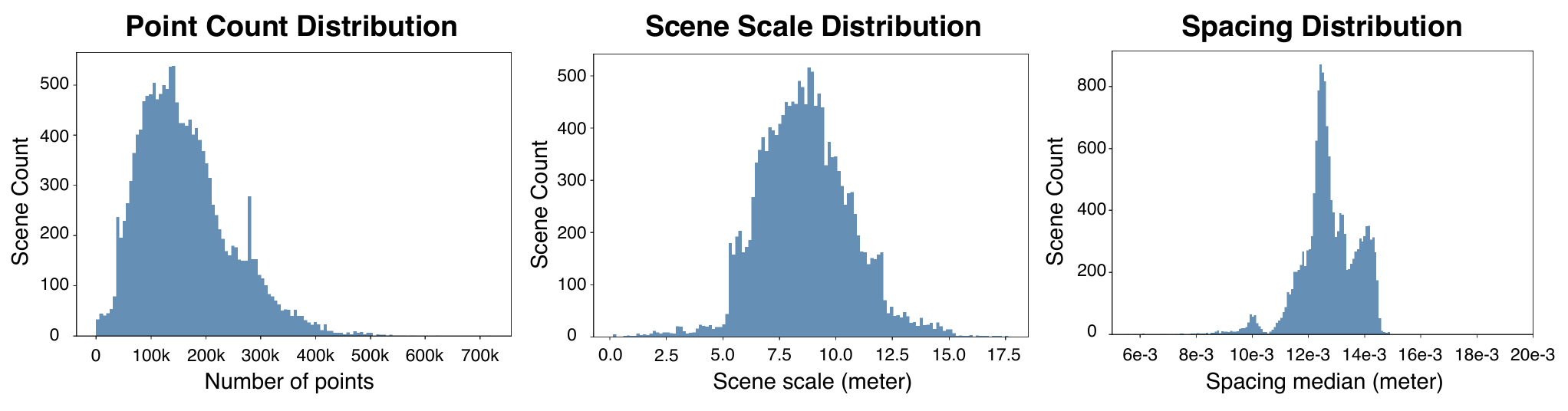}
    \caption{\textbf{Statistics of the video-generated point clouds.} We report the properties of the aligned scenes used for pre-training. (Left) Distribution of the number of points per scene before the align processing.  (Center) Distribution of scene scales (the diagonal length of the axis-aligned bounding box). (Right) Distribution of the median point spacing, defined as the median kNN distance within each scene.}
    \label{fig:scene_statistics}
\end{center}%
}]

\section{\data{} Details}
We construct \data using a feedforward reconstruction model applied to unlabeled videos from the web.  The dataset consists of VGPC obtained from both our collected videos and existing datasets (RealEstate10k~\cite{zhou2018stereo}, YouTube House Tours~\cite{chang2020semantic}, and HouseTours~\cite{Celen_2025_ICCV}). In this section, we report statistics of our independently collected videos and the VGPC reconstructed from them.

For our own collection, we assume that indoor layouts and furniture vary across geographic regions.  Therefore, we collected walkthrough videos from 19 countries, as shown in Fig.~\ref{fig:collect_video}. In total, we gathered 3,462 videos, each with a median duration of 3.63 minutes.

For scene classification, we sample every frame at the native FPS of each video and resize all frames to the standard CLIP~\cite{radford2021learning} ViT-B/32 resolution of $224 \times 224$. Each frame is first classified by the CLIP image encoder as either “indoor’’ or “outdoor.’’  Frames predicted as indoor are then re-evaluated using room-type prompts (living room, bedroom, bathroom) to obtain scene categorization. 
This two-stage filtering produces a total of 15,921 indoor sequences, with each video contributing on average 4.59 sequences.

The reconstruction of a single scene takes approximately 5 minutes on average, although the actual time depends on the number of frames extracted from each video.  We process our collected videos using eight NVIDIA H200 GPUs (144GB each), requiring roughly eight hours in total to generate the video-based point clouds. The raw outputs are pixel-aligned point clouds and therefore contain an extremely large number of points. Directly applying our post-processing pipeline to these raw point clouds leads to substantial computational overhead. To maintain efficiency, we randomly downsample each scene to 20k points before alignment and normalization. Fig.~\ref{fig:scene_statistics} reports the statistics of the post-processed VGPC, including the distribution of point counts and scene scale. After post-processing, the structural properties of the indoor scenes closely match those observed in standard ScanNet scenes.

\begin{table}[t]
\centering
\caption{Configuration of PTv3 (Base) and PTv3 (Large)}
\vspace{-10pt}
\resizebox{\linewidth}{!}{
\begin{tabular}{l c c}
\toprule
\multirow{2}{*}{Config} & \multicolumn{2}{c}{Value} \\
\cmidrule(lr){2-3}
 & PTv3 (Base) & PTv3 (Large) \\
\midrule
order & Z + TZ + H + TH 
      & Z + TZ + H + TH  \\
stride & (2, 2, 2, 2) & (2, 2, 2, 2) \\
enc\_depths & (3, 3, 3, 12, 3) & (3, 3, 3, 12, 3) \\
enc\_channels & (48, 96, 192, 384, 512) & (64, 128, 256, 512, 768) \\
enc\_num\_head & (3, 6, 12, 24, 32) & (4, 8, 16, 32, 48) \\
enc\_patch\_size & 1024 × 5 & 1024 × 5 \\
mlp\_ratio & 4 & 4 \\
qkv\_bias & True & True \\
drop\_path & 0.3 & 0.3 \\
head\_in\_channels & 1088 & 1536 \\
head\_embed\_channels & 256 & 256 \\
head\_prototypes & 4096 & 4096 \\
params & 121M & 224M \\
\bottomrule
\end{tabular}
}
\label{tab:ptv3_config}
\vspace{-5pt}
\end{table}

\begin{table}[t]
\centering
\caption{Pre-training configuration for \method{}}
\vspace{-10pt}
\resizebox{0.99\linewidth}{!}{
\begin{tabular}{l c}
\toprule
Config & Value \\
\midrule
laplacian kNN & 24 \\
huber delta & 0.5 \\
max radius distance & 0.08 \\
laplacian loss weight & 2e-4 $\rightarrow$ 3e-3 \\
noise consistency loss weight & 0.05 \\
teacher temperature & 0.04 $\rightarrow$ 0.07 \\
student temperature & 0.1 \\
views (global / local) & 2 / 4 \\
optimizer & AdamW \\
base learning rate & 1e-3 \\
layer-wise LR decay & 0.9 \\
weight decay & 0.04 $\rightarrow$ 0.10 \\
scheduler & OneCycleLR \\
batch size & 16 \\
& 145,600 (\data-1k, PTv3-Base) \\ 
iterations & 291,200 (\data-49k, PTv3-Base)\\ 
& 436,800 (\data-49k, PTv3-Large) \\
\bottomrule
\end{tabular}
}
\label{tab:lam3c_config}
\vspace{-10pt}
\end{table}

\section{Implementation Details}
This section explains the details of our implementation. We begin by describing Point Transformer V3 (PTv3)~\cite{Wu_2024_CVPR} which is the backbone model used in our experiments. We then detail the pre-training setup and the configurations adopted for semantic segmentation and instance segmentation. 
Our implementation is based on PointCept~\cite{pointcept2023}.

\noindent{\textbf{Backbone network.}}
PTv3 improves efficiency by removing the kNN query and the Relative Positional Encoding (RPE) modules in PTv2~\cite{wu2022point}. The kNN query module was designed to capture local structure on unstructured point clouds, but it requires repeated distance computations at every iteration and occupies 28\% of the forward time. The RPE module encodes relative positions, yet point cloud RPE must compute pairwise Euclidean distances for every point pair, accounting for 26\% of the computation. PTv3 avoids both costs by abandoning the unordered-set formulation and serializing point clouds into a structured representation, which eliminates the need for kNN and RPE computations and leads to efficiency gains. PTv3 assigns a consistent order to points through four ordering strategies: Z-order (Z), Transposed Z-order (TZ), Hilbert (H), and Transposed Hilbert (TH). 

We use PTv3 as the backbone model for all experiments, with two variants: PTv3 (Base) and PTv3 (Large). The Base variant follows the configuration used in Sonata, and the Large variant follows the configuration used in Concerto~\cite{zhang2025concerto}. The detailed parameter settings are provided in Table~\ref{tab:ptv3_config}.

\begin{table}[t]
\centering
\caption{Indoor semantic segmentation settings}
\vspace{-10pt}
\resizebox{\linewidth}{!}{
\begin{tabular}{lclc}
\toprule
\multicolumn{2}{c}{Linear Probing} &\multicolumn{2}{c}{Full Fine-Tuning} \\
\cmidrule(lr){1-2} \cmidrule(lr){3-4}
Config &Value &Config &Value \\
\midrule
optimizer & AdamW & optimizer & AdamW \\
scheduler & OneCycleLR & scheduler & OneCycleLR \\
criteria & CrossEntropy & criteria & CrossEntropy \\
learning rate & 2e-3 & learning rate & 2e-3 \\
block LR & N/A & block LR & 2e-4 \\
weight decay & 2e-2 & weight decay & 2e-2 \\
batch size & 168 & batch size & 128 \\
epochs & 100 & epochs & 100 \\
\bottomrule
\end{tabular}
}
\label{tab:semantic_config}
\vspace{-5pt}
\end{table}

\begin{table}[t]
\centering
\caption{Indoor instance segmentation settings}
\vspace{-10pt}
\resizebox{\linewidth}{!}{
\begin{tabular}{lclc}
\toprule
\multicolumn{2}{c}{Linear Probing} & \multicolumn{2}{c}{Full Fine-Tuning} \\
\cmidrule(lr){1-2} \cmidrule(lr){3-4}
Config & Value & Config & Value \\
\midrule
optimizer & AdamW & optimizer & AdamW \\
scheduler & OneCycleLR & scheduler & OneCycleLR \\
criteria & CrossEntropy & criteria & CrossEntropy \\
learning rate & 2e-3 & learning rate & 2e-3 \\
block LR & N/A & block LR & 2e-4 \\
weight decay & 5e-2 & weight decay & 5e-2 \\
batch size & 168 & batch size &
\begin{tabular}{@{}c@{}}
168 (ScanNet / S3DIS)\\
144 (ScanNet200)\\
96 (ScanNet++)
\end{tabular} \\
epochs & 100 & epochs & 100 \\
\bottomrule
\end{tabular}
}
\label{tab:instance_config}
\vspace{-10pt}
\end{table}

\begin{table*}[t]
\vspace{-.2em}
\centering

\subfloat[
\textbf{kNN neighborhood size}
\label{tab:knn}
]{
\begin{minipage}[t]{0.31\linewidth}\vspace{0pt}
\begin{center}
\tablestyle{4pt}{1.05}
\begin{tabular}{x{48}x{32}x{32}}
$k$ & LP & Full-FT \\
\shline
8  & 56.0 & \textbf{75.9} \\
24   & \baseline{\textbf{57.1}} & \baseline{75.1} \\
32   & 54.9 & 75.2 \\
\end{tabular}
\end{center}
\end{minipage}
}
\hspace{2em}
\subfloat[
\textbf{Radius threshold for noisy point removal.} 
\label{tab:radius}
]{
\begin{minipage}[t]{0.31\linewidth}\vspace{0pt}
\begin{center}
\tablestyle{4pt}{1.05}
\begin{tabular}{x{48}x{32}x{32}}
radius & LP & Full-FT \\
\shline
0.04  & 55.3 & \textbf{75.7} \\
0.08   & \baseline{\textbf{57.1}} & \baseline{75.1} \\
0.12   & 55.3 & 75.4 \\
\end{tabular}
\end{center}
\end{minipage}
}

\par\vspace{0.4em}


\subfloat[
\textbf{Huber loss parameter $\delta$}. 
\label{tab:huber_loss}
]{
\begin{minipage}[t]{0.29\linewidth}\vspace{0pt}
\begin{center}
\tablestyle{4pt}{1.05}
\begin{tabular}{x{64}x{32}x{32}}
$\delta$ & LP & Full-FT \\
\shline
0.05  & 55.7 & \textbf{75.5} \\
0.5   & \baseline{\textbf{57.1}} & \baseline{75.1} \\
1.0   & 55.4 & 75.3 \\
\end{tabular}
\end{center}
\end{minipage}
}
\hspace{1.2em}
\subfloat[
\textbf{Gaussian scale parameter $\sigma$.} 
\label{tab:sigma}
]{
\begin{minipage}[t]{0.29\linewidth}\vspace{0pt}
\begin{center}
\tablestyle{4pt}{1.05}
\begin{tabular}{x{72}x{32}x{32}}
$\sigma$ & LP & Full-FT \\
\shline
adaptive &  \baseline{\textbf{57.1}} & \baseline{75.1}\\
0.03   & 56.0 & \textbf{75.4} \\
0.08   & 56.2 & 75.0 \\
\end{tabular}
\end{center}
\end{minipage}
}
\hspace{1.2em}
\subfloat[
\textbf{Loss Weights $\lambda$ and $\mu$.}
\label{tab:loss_weight}
]{
\begin{minipage}[t]{0.29\linewidth}\vspace{0pt}
\begin{center}
\tablestyle{4pt}{1.05}
\begin{tabular}{x{18}x{18}x{18}x{32}x{32}}
$\lambda_{\text{start}}$ & $\lambda_{\text{base}}$ & $\mu$ & LP & Full-FT \\
\shline
2e-4 & 3e-3 & 5e-2 & \baseline{\textbf{57.1}} & \baseline{75.1} \\
2e-4 & 3e-4 & 2e-2 & 56.7 & \textbf{75.7} \\
2e-4 & 3e-4 & 5e-2 & 56.5 & 75.2 \\
2e-4 & 3e-3 & 2e-2 & 54.9 & 74.8 \\
\end{tabular}
\end{center}
\end{minipage}
}

\vspace{-.1em}
\caption{\textbf{Hyperparameter Analysis Experiments} on the ScanNet semantic segmentation. We report Full Fine-tuning (Full-FT) and Linear Probing (LP) performance (\%).  Unless noted otherwise, the default configuration uses \data{}-1k generated by $\pi^3$ and is pre-trained with PTv3 (Base).  Both Full-FT and LP are trained for 100 epochs. Default settings are highlighted in \colorbox{baselinecolor}{gray}.}
\label{tab:ablations}
\vspace{-10pt}
\end{table*}

\noindent{\textbf{Pre-training setting.}}
We set the default parameters as shown in Table~\ref{tab:lam3c_config}. The Laplacian smoothing loss hyperparameters were set based on experimental results and computational considerations during pre-training. 
In our formulation, the Laplacian smoothing loss penalizes pairwise feature differences between neighboring points on the kNN graph. When the feature difference on an edge is small, the Huber penalty behaves quadratically; when it exceeds $\delta$, it becomes linear, reducing the influence of outlier edges.
Other settings related to the clustering loss follow the configuration used in Sonata. Furthermore, our comparative experiments employed three settings: RoomTours-16k (PTv3-Base), RoomTours-49k (PTv3-Base), and RoomTours-49k (PTv3-Large). Based on empirical results from previous research, the number of iterations is scaled with both data size and model size.

\noindent{\textbf{Downstream task setting.}}
We evaluate the pre-training effect on two downstream tasks: semantic segmentation and instance segmentation in indoor scenes. Semantic segmentation and instance segmentation are evaluated on ScanNet~\cite{dai2017scannet}, ScanNet200~\cite{rozenberszki2022language}, ScanNet++~\cite{yeshwanth2023scannet++}, S3DIS~\cite{Armeni_2016_CVPR}. For both tasks, we adopt the following fine-tuning protocols: Linear probing and Full fine-tuning. Linear probing freezes the entire backbone and trains only a linear classifier, whereas Full fine-tuning updates all model parameters. For the linear probing and the full fine-tuning protocol, we follow Sonata setting~\cite{wu2025sonata}. The configuration for semantic segmentation is provided in Table~\ref{tab:semantic_config}, and the configuration for instance segmentation is provided in Table~\ref{tab:instance_config}.

\section{Further Analysis Experiments}
We analyze the key hyperparameters of \method{}.
Our ablations focus on the neighborhood size for Laplacian smoothing loss, the radius threshold for noisy point removal, the Huber loss parameter $\delta$, the Gaussian scale parameter $\sigma$, and the loss weights $\lambda$ and $\mu$ for Laplacian smoothing and noise consistency terms.

\noindent{\textbf{kNN neighborhood size (see Table~\ref{tab:knn}).}}
We analyze the effect of the neighborhood size used to construct the kNN graph for the Laplacian smoothing loss. In our \method{}, the Laplacian smoothing loss is computed on a kNN graph constructed for each VGPC. The value of $k$ determines the size of the local neighborhood, thus controlling the strength and spatial extent of the smoothing imposed by the regularization.

We pre-train \method{} with three values $k {=} \{8, 24, 32\}$, and evaluate the resulting models on ScanNet semantic segmentation.
As shown in Table~\ref{tab:knn}, $k {=} 24$ yields the best performance, particularly in the Linear Probing setting where representation quality is most directly reflected.
We consider that smaller $k$ results in unstable local neighborhoods, while larger $k$ leads to over-smoothing.

\noindent{\textbf{Radius threshold for noisy point removal (see Table~\ref{tab:radius}).}}
We analyze the effect of the radius threshold used to remove distant neighbors when constructing the kNN graph for the Laplacian smoothing loss. In VGPC, a noisy reconstructed point may cause a query point to retrieve far-away points among its top-k neighbors, which violates the locality assumption required for Laplacian regularization. To mitigate this issue, we apply a distance threshold  $r_{max}$ and discard neighbors whose Euclidean distance exceeds this value.

We evaluate three thresholds $r {=} \{0.04, 0.08, 0.12\}$, by pre-training \method{} with each setting and conducting semantic segmentation on ScanNet. As shown in Table~\ref{tab:radius}, $r{=}0.08$ achieves the best performance, particularly in the Linear Probing setting, indicating that this threshold strikes an effective balance between removing noisy outliers and preserving sufficient local structure.

\begin{table*}[t]
\centering
\caption{\textbf{Estimating the effect of test/val set contamination.} All self-supervised methods are evaluated by full fine-tuning (Full-FT) or linear probing (LP) on PTv3 (Base) for 100 epochs. We use mIoU as evaluation metric.}
\vspace{-10pt}
\label{tab:testval_eval}
\scalebox{0.97}{%
\begin{tabular}{p{28.5mm}rp{7.5mm}ccccccccccc}
\toprule
Semantic Seg. &\multicolumn{2}{c}{Test/Val set} &\multicolumn{2}{c}{ScanNet~\cite{dai2017scannet}} &\multicolumn{2}{c}{ScanNet200~\cite{rozenberszki2022language}} &\multicolumn{2}{c}{ScanNet++ Val~\cite{yeshwanth2023scannet++}} &\multicolumn{2}{c}{S3DIS Area 5~\cite{Armeni_2016_CVPR}} \\
\cmidrule(lr){1-1} \cmidrule(lr){2-3} \cmidrule(lr){4-5} \cmidrule(lr){6-7} \cmidrule(lr){8-9} \cmidrule(lr){10-11}
Methods &\multicolumn{1}{c}{With} &\multicolumn{1}{c}{Without} & LP & Full-FT & LP & Full-FT & LP & Full-FT & LP & Full-FT  \\
\midrule
  Sonata (ScanNet)  & \multicolumn{1}{c}{\cmark} & \multicolumn{1}{c}{--}
 & \textbf{67.3} & \textbf{77.4 }
 & 26.6 & \textbf{32.7}
 & \textbf{35.1} & \textbf{42.4} 
 & \textbf{63.4} & \textbf{72.5}
 \\

  Sonata (ScanNet) & \multicolumn{1}{c}{--} & \multicolumn{1}{c}{\cmark}
 & 67.1 & 75.4 
 & \textbf{27.2} & 32.2
 & 34.3 & 41.7 
 & 61.2 & 72.2  
 \\
\bottomrule
\end{tabular}
}
\end{table*}

\noindent{\textbf{Huber loss parameter $\delta$ (see Table~\ref{tab:huber_loss}).}}
We analyze the effect of the Huber loss parameter $\delta$, which determines the transition point between the quadratic and linear regimes of the Huber loss.
In our formulation, the Laplacian smoothing loss penalizes pairwise feature differences between neighboring points on the kNN graph. When the feature difference on an edge is small, the Huber penalty behaves quadratically; when it exceeds $\delta$, it becomes linear, reducing the influence of outlier edges.

We evaluate three settings $\delta {=} \{0.05, 0.5, 1.0\}$ and assess the pre-trained models on ScanNet semantic segmentation. 
As shown in Table~\ref{tab:huber_loss}, $\delta {=} 0.5$ achieves the best Linear Probing performance, suggesting that it provides an effective balance between sensitivity to local variations and robustness to noisy neighborhoods.

\noindent{\textbf{Gaussian scale parameter $\sigma$ (see Table~\ref{tab:sigma}).}}
We analyze the effect of the Gaussian scale parameter $\sigma$, which controls the decay rate of the distance-based weighting used in the Laplacian smoothing loss. 
We assign each edge a distance-based weight \(
w_{ij} = \exp(-{\| p_i - p_j \|^2}/{\sigma^2})
\), where a larger $\sigma$ results in a slower decay of the weights, allowing distant points to retain non-negligible influence. Conversely, a smaller $\sigma$ produces a sharper decay, restricting the effective neighborhood to only very close points. To account for the varying density of VGPC, our method adaptively sets $\sigma$ as the median of the kNN distances for each point cloud. 

We compare three settings, $\sigma {=} \{\text{adaptive}, 0.03, 0.08\}$, and evaluate the pre-trained models on ScanNet semantic segmentation. As shown in Table~\ref{tab:sigma}, the adaptive setting achieves the best performance, particularly in the Linear Probing evaluation, indicating that adaptive scaling provides robustness across point clouds with diverse geometric densities.

\noindent{\textbf{Loss weights $\lambda$ and $\mu$ (see Table~\ref{tab:loss_weight}).}}
We analyze the impact of the weighting coefficients $\lambda$ and $\mu$ used for the Laplacian smoothing loss and Noise consistency loss, respectively. 
As defined in Eq.~(5), the overall objective consists of the distillation loss from Sonata and two additional regularization terms. Here, we focus on the balance between these two regularized losses, which are introduced to stabilize representation learning from VGPC. In our formulation, the coefficient $\lambda$ for the Laplacian smoothing loss follows a scheduled progression during training, whereas the coefficient $\mu$ for the Noise consistency loss is fixed. 

We evaluate four combinations of $(\lambda, \mu)$ listed in Table~\ref{tab:loss_weight}, pre-train \method{} under each setting, and assess the resulting models on ScanNet semantic segmentation.
As shown in Table~\ref{tab:loss_weight}, the configuration 
$\lambda_{\text{start}} {=} 2\times10^{-4}$, 
$\lambda_{\text{base}} {=} 3\times10^{-3}$, 
and $\mu = 5\times10^{-2}$ 
achieves the best Linear Probing performance, indicating that this balance of regularization strengths leads to more stable and discriminative representations.

\begin{figure}[t]
  \centering
  \includegraphics[width=\linewidth]{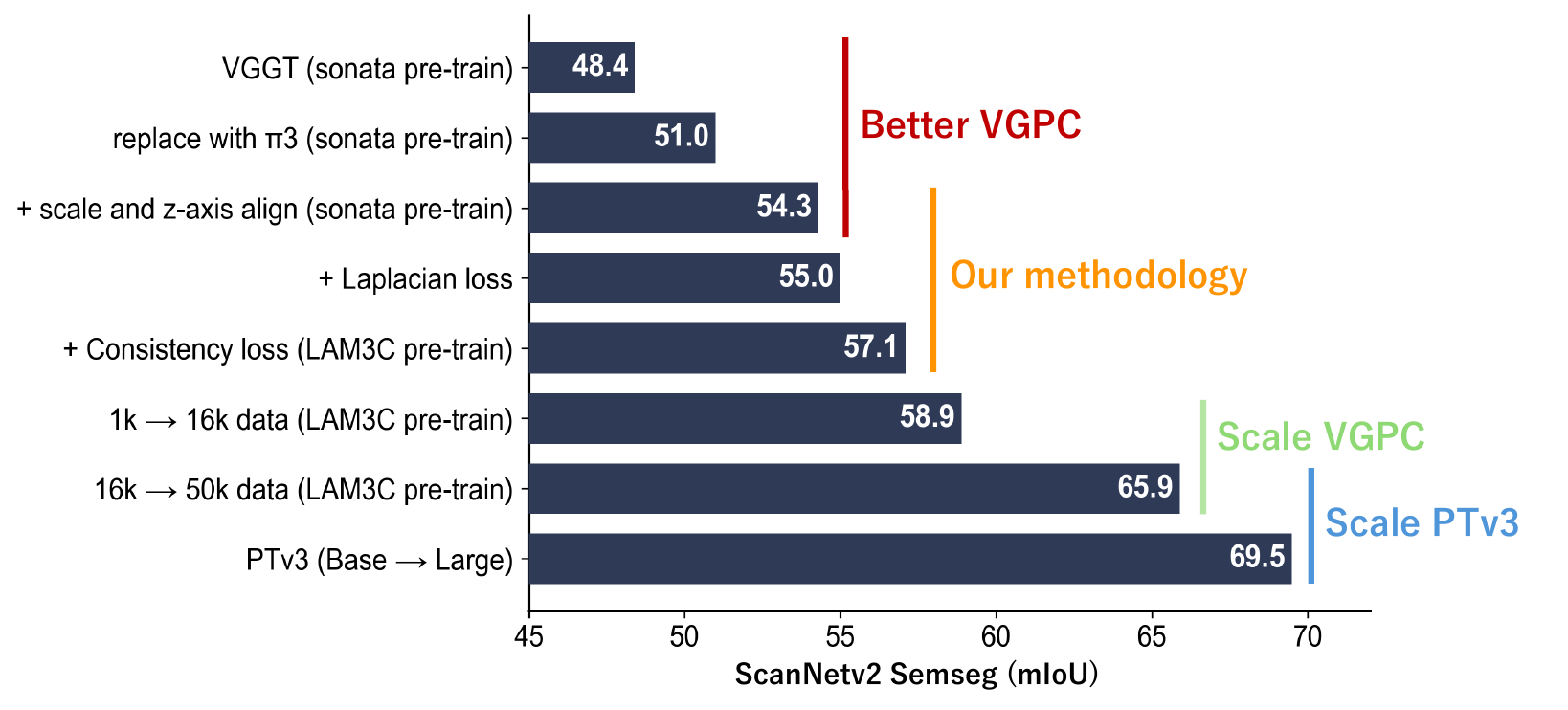}
  \caption{Component synergy on ScanNet semantic segmentation (linear probing). Replacing VGGT with $pi^3$, adding indoor alignment, and introducing LAM3C each improve performance, with further gains from scaling RoomTours and using PTv3-Large.}
  \label{fig:ablation}
\end{figure}

\noindent{\textbf{Component synergy (see Figure~\ref{fig:ablation}).}}
We add a combination analysis on ScanNet linear probing. Keeping the same Sonata pre-training, switching the reconstruction model (VGGT→$\pi^3$), adding indoor alignment, and further adding \method consistently improve performance, with continued gains when increasing data scale and model capacity, indicating complementary and synergistic effects between components.

\section{Fairness considerations for baseline}

In the paper, we compare against Sonata as it is the current state of the art. However, the official Sonata models are pre-trained on all splits of ScanNet, ScanNet++, and S3DIS including the \{train / val / test\} splits used for downstream evaluation\footnote{The official configuration can be verified in the Sonata GitHub repository: \\
\url{https://github.com/Pointcept/Pointcept/blob/main/configs/sonata/pretrain-sonata-v1m1-0-base.py}}. 
This setup is not comparable to our \method{}, which, even during finetuning evaluations, does not see any  \{test / val\}  data. 
Such \{test / val\} inclusion has been common practice in previous indoor 3D-SSL methods~\cite{wu2023masked,wu2024towards}, but blurs the true generalisation performance measurement. 

To ensure a fair comparison, we construct a fair Sonata baseline by pre-training Sonata using only the train split of ScanNet, completely removing the corresponding \{val / test\} scenes from pre-training.

We first examine how excluding the \{val / test\} scenes affects Sonata's performance when pre-trained on ScanNet. Table~\ref{tab:testval_eval} reports the results for semantic segmentation. 
Removing \{val / test\} leads to a consistent drop in performance, confirming that the standard Sonata configuration benefits from the exposure to the evaluation scenes. 
This leakage provides prior knowledge of scene and object layouts, leading to performance that cannot be attributed purely to generalization. 
Although this paper focuses on ScanNet, we expect the performance gap to widen when combining multiple datasets such as ScanNet++, S3DIS, and others, where leakage accumulates across datasets.

These findings highlight a fundamental issue in how 3D-SSL methods have been evaluated. 
While a deeper investigation is beyond the scope of this paper, we ensure fairness in all experiments by using Sonata models pre-trained strictly on the train split only.

\section{Discussion}
Our findings show that high-fidelity 3D scans are not a prerequisite for effective 3D-SSL. Despite being noisy, incomplete, and inconsistent, VGPC reconstructed from unlabeled videos prove sufficient for 3D-SSL. Perhaps most notably, \method{} performs strongly in linear probing, a setting that has not been widely explored in previous 3D-SSL methods. This indicates that these imperfect reconstructions still encode rich geometric cues relevant to real-world tasks. Overall, our results highlight a promising direction toward scalable 3D data generation. Instead of costly real 3D scans, unlabeled videos offer a far more scalable source of 3D supervision, where the key is learning to handle noisy and missing regions effectively.

\section{Limitations and outlook} 
Although the reconstruction models used in this paper use real point clouds during training, our core contribution lies in showing that previously untapped resources such as web videos can be used for 3D-SSL and become more useful at scale. By piggybacking on reconstruction models, our approach benefits from the fast progress in this domain and might enable unifying 3D reconstruction and recognition models. Our method is designed for predominantly static indoor scenes and may be limited in highly dynamic environments, where reconstruction quality degrades due to many moving objects. While our noise regularization mitigates some outliers in \data, explicit modeling of dense dynamics and motion remains an important direction for future work.


\end{document}